\documentclass{article}

\usepackage[final]{neurips_2024}

\usepackage[utf8]{inputenc}       
\usepackage[T1]{fontenc}          
\usepackage{hyperref}             
\usepackage{url}                  
\usepackage{booktabs}             
\usepackage{amsfonts}             
\usepackage{nicefrac}             
\usepackage{microtype}            
\usepackage[table]{xcolor}               

\usepackage{wrapfig}              
\usepackage{float}                
\usepackage{graphicx}             
\usepackage{amsmath}              
\usepackage{comment}              
\usepackage[capitalize]{cleveref} 

\usepackage{acro}                
\usepackage{caption}             
\usepackage{subcaption}          
\usepackage{placeins}            

\usepackage[breakable]{tcolorbox} 
\usepackage[dvipsnames]{xcolor}   

\newcommand{\uncertainty}[2]{%
    #1 {\scalebox{0.8}{\textcolor{gray}{$\pm$ #2}}}%
}

\usepackage[shortlabels]{enumitem}

\usepackage{tablefootnote}

\setcitestyle{authoryear,round,citesep={;},aysep={,},yysep={;}}

\definecolor{mydarkblue}{rgb}{0,0.08,0.45}
\hypersetup{ %
    pdftitle={},
    pdfsubject={AI4Mat Workshop at International Conference on Machine Learning 2024},
    pdfkeywords={},
    pdfborder=0 0 0,
    pdfpagemode=UseNone,
    colorlinks=true,
    linkcolor=mydarkblue,
    citecolor=mydarkblue,
    filecolor=mydarkblue,
    urlcolor=mydarkblue,
    }

\DeclareAcronym{rl}{
  short=RL,
  long=Reinforcement Learning,
}
\title{AtomComposer: Discovering Chemical Space from First Principles with Reinforcement Learning}

\author{%
Bjarke Hastrup\textsuperscript{1}, 
François Cornet\textsuperscript{1,2}, 
Tejs Vegge\textsuperscript{1,3}, 
Arghya Bhowmik\textsuperscript{1,3}\thanks{Corresponding author: \texttt{arbh@dtu.dk}} \\
\texttt{\{bjaha, frjc, teve, arbh\}@dtu.dk} \vspace{0.4em} \\
\textsuperscript{1} Dept. of Energy Conversion and Storage, Technical University of Denmark, Denmark \\
\textsuperscript{2} Dept. of Applied Mathematics and Computer Science, Technical University of Denmark, Denmark \\
\textsuperscript{3} Pioneer Center for Accelerating P2X Materials Discovery (CAPeX), Kgs. Lyngby, Denmark
}
\begin{document}
\maketitle


\begin{abstract}

Discovering novel stable molecules without training data remains a grand scientific challenge. Current molecular generative models are trained on large, pre-curated datasets, which introduce biases and limit exploration of novel chemistry. In contrast, we propose a new paradigm: autonomous, generalized agents capable of mapping vast, unknown chemical spaces without any pretraining. 
For the first time, we present AtomComposer, a self-guided agent that autonomously constructs valid 3D isomers under stoichiometric constraints and is trained exclusively online using reinforcement learning. Unlike existing approaches that generally overfit to a specific chemical formula, we establish a multi-composition training scheme that enables a broad generalization across diverse chemistry, guided by energy- and validity-based rewards. Our agent can discover up to an order of magnitude more valid isomers on unseen test formulas than existing single-composition reinforcement-learning baselines trained with per-step energy rewards. These results fulfill the promise of online reinforcement learning as a powerful paradigm for scalable, from-scratch exploration of chemical configuration space. 
\end{abstract}

\section{Introduction}

Autonomous discovery of novel molecules with bespoke properties is the new frontier in computational chemistry. Effectively navigating the vast chemical space requires innovative and data-efficient search strategies, and has motivated the development of a broad range of methodologies including quantum-mechanical property-guided machine learning, inverse molecular design, and automated chemical space exploration frameworks~\citep{von2020exploring, kalikadien2022chemspax, ioannidis2016molsimplify, khalak2022chemical, sandonas2023freedom, fallani2024inverse}. Generative models have emerged as a particular promising avenue for this task~\citep{anstine2023generative}, yet their performance often hinges on the availability of suitable training data. Large public datasets are rarely curated with specific property optimization in mind, and the relevant property regimes may lie at the fringes---or entirely outside---the observed data distribution~\citep{brown2019guacamol}. This poses a fundamental challenge: models must not only interpolate but also extrapolate beyond known examples~\citep{schrier2023pursuit}. Constructing task-specific datasets is likewise non-trivial and, even when feasible, introduces chemical and structural biases that may limit exploration of novel chemical spaces.


Reinforcement learning (RL)~\citep{sutton2018reinforcement} is a computational framework in which an autonomous agent learns to make sequential decisions by interacting with an environment and receiving feedback in the form of rewards. Rather than learning from a fixed dataset of examples, the agent improves its behavior through trial and error, gradually discovering strategies that maximize cumulative reward. In the context of molecular discovery, the agent can be thought of as a chemist constructing molecules atom by atom, guided not by prior knowledge of what good molecules look like, but purely by evaluating the quality of the structures it produces. This makes RL particularly well suited to exploring regions of chemical space that are poorly represented in existing datasets, where data-driven generative models would struggle to extrapolate reliably~\citep{sridharan2024deep}.

Beyond mitigating dataset-induced bias, \textit{tabula rasa} RL (learning entirely from scratch without pretraining on existing molecular data) is well aligned with the sequential and long-horizon nature of atom-by-atom molecular construction. In this setting, chemically meaningful objectives such as stability and chemical validity are only available once a full 3D structure has been assembled, resulting in sparse and delayed reward signals~\citep{sutton2018reinforcement}.

Classical and more recent online structure-search approaches—including genetic algorithms, basin-hopping, minima-hopping, and their machine-learning-accelerated variants—have been successfully applied across a wide range of molecular, cluster, and materials discovery problems~\citep{johnston2003evolving, jensen2019graph, wales1997global, goedecker2004minima, meldgaard2018machine, christiansen2022atomistic}. These methods are typically formulated as stochastic global optimizers, in which domain knowledge is encoded through fixed proposal mechanisms (e.g., mutation, crossover, or displacement moves), while the search itself does not adapt through learned, parametric decision policies.
In this sense, and analogous to deep generative models whose outputs reflect biases in their training data, these approaches—and more broadly human intuition-driven motif design—are constrained by existing chemical knowledge. More generally, such biases arise from the design of the generative process itself (e.g., choice of move sets or construction rules), which implicitly defines the accessible search space.
In contrast, reinforcement learning explicitly learns a construction policy from online interaction, enabling a less prescriptive mode of exploration in which structure-building strategies are discovered rather than fixed a priori. 
This allows chemical design principles to be represented, updated, and reused across tasks, thereby offering a complementary paradigm for composition-generalizable molecular discovery~\citep{schulman2017proximal, jorgensen2019atomistic, mortensen2020atomistic, meldgaard2020structure, molgym1}.
However, although the learning policy can flexibly adapt within this space, the underlying construction framework (\textit{action space} in RL terminology) still imposes inductive biases, such that the degree of open-ended exploration ultimately depends on the expressiveness of this construction formulation.




A substantial body of RL research has employed two alternative molecular representations: SMILES strings and 2D molecular graphs~\citep{weininger1988smiles}. Pioneered by \cite{olivecrona2017molecular} and \cite{popova2018deep} and further developed in REINVENT 2.0~\citep{blaschke2020reinvent}, SMILES-based RL agents learn to generate molecules as character sequences and can be efficiently fine-tuned toward diverse property objectives. Graph-based RL approaches such as MolDQN~\citep{zhou2019optimization} and GCPN~\citep{you2018graph} instead operate directly on molecular connectivity graphs, adding or modifying atoms and bonds as discrete actions. Both SMILES and graph representations encode molecular topology—which atoms are bonded to which—but omit explicit 3D geometry~\citep{elton2019deep}. As a consequence, a single SMILES string or connectivity graph may correspond to multiple distinct three-dimensional structures: different conformers arising from rotatable bonds, as well as stereoisomers (including enantiomers and diastereomers) arising from chiral centers or restricted rotation~\citep{hawkins2017conformation}.

This representational ambiguity has important practical consequences: a conformational search must be performed post hoc to enumerate the relevant candidate geometries and evaluate their relative stabilities. For larger molecules or those with multiple chiral centers, the number of distinct structures grows combinatorially, making exhaustive conformer search a challenging problem in its own right~\citep{leach2007introduction}. Automated conformer generators such as ETKDG~\citep{riniker2015better} provide a practical approximation, but tend to undersample higher-energy conformers and may miss the most stable or most functionally relevant geometry entirely~\citep{friedrich2017high, hawkins2017conformation}. Since many molecular properties depend sensitively on shape~\citep{von2020exploring}, generating molecules directly in 3D avoids this post-processing bottleneck and ensures that property evaluations are grounded in explicit, physically meaningful atomic coordinates from the outset.

In the supervised setting, some of the most promising directions for molecule generation in 3D are either based on denoising diffusion~\citep{hoogeboom2022equivariant,le2024navigating,cornet2024equivariant}, flow matching~\citep{song2023equivariant,irwin2025semlaflow}, or auto-regressive models that build molecules in an fragment-by-fragment or atom-by-atom fashion~\citep{NEURIPS2019_a4d8e2a7, gebauer2022inverse, roney2022generating,
daigavane2023symphony,ochoa2024molminer}. Although diffusion models could potentially be integrated into a pretraining-finetuning framework~\citep{black2024training}, it remains unclear whether they can effectively be used for \textit{tabula rasa} learning for navigating a $3$N dimensional energy surface.
In the purely online setup, RL has been used for conformer~\citep{jiang2022conformer, D4DD00023D} and isomer~\citep{molgym1,molgym2} generation.~\cite{flam2022scalable} extended \textsc{MolGym} to place fragments instead of individual atoms, improving scalability and the size of the generated molecules.~\cite{meldgaard2021generating} used online RL but only after an offline pretraining phase. Whereas their pretraining was multi-compositional, their online finetuning was for single compositions only and further relied on result aggregation from 64 parallel fine-tunings spawned after pertaining. As a result, a general and scalable demonstration of \ac{rl} for \textit{tabula rasa} $3$D molecular structure discovery remains lacking. Existing results show only limited success on simple organic molecules~\citep{molgym1, molgym2, meldgaard2021generating} or metal clusters of fixed composition~\citep{elsborg2023equivariant, modee2023megen}, with poor generalization across stoichiometries. A key reason is that most prior approaches rely on training regimes with limited compositional diversity, in which the policy and value function are optimized for a single fixed composition at a time. This leads to overfitting to composition-specific construction strategies rather than policies that transfer across chemical space.



The evaluation of RL algorithms for molecular structure discovery presents a fundamental challenge. In the standard generative modeling paradigm, where models are trained via supervised learning to approximate the training distribution, performance is typically assessed through stochastic rollouts at a single final checkpoint, reflecting how well the converged model captures the underlying data~\citep{gomez2018automatic, NEURIPS2019_a4d8e2a7, cornet2024om}. In contrast, performance evaluation in RL is considerably more nuanced~\citep{henderson2018deep, dulac2020empirical}. The performance of an RL agent can vary significantly depending on the checkpoint selected, as the underlying policy evolves throughout training~\citep{islam2017reproducibility}. A single-checkpoint evaluation, though convenient and widely used~\citep{xia2022systematic}, often fails to capture the broader behavioral dynamics and exploration strategies adopted at different training stages~\citep{colas2019hitchhiker}. This evolving behavior, combined with the open-ended nature of RL tasks, renders traditional metrics, such as distance to a reference dataset, largely inadequate. Instead, a meaningful evaluation often requires the direct involvement of a chemist to quantify the utility of individually generated structures through score/reward functions~\citep{brown2019guacamol, polykovskiy2020molecular, schwalbe2020generative}. This reliance on task-specific metrics complicates automation and has hindered the establishment of universally recognized benchmark tasks, making it difficult to objectively compare algorithms and slowing methodological convergence in the field~\citep{olivecrona2017molecular, xie2021mars, nie2024durian}.


A core challenge in online RL for \textit{$3$D molecular} discovery in particular, is balancing the delicate trade-off between exploration and physical stability. While policy stochasticity is essential for escaping locally optimal behavior~\citep{haarnoja2018soft, schulman2017proximal}, excessive spatial noise can corrupt energy based evaluations, which depend sensitively on the atomic coordinates~\citep{smith2017ani, gastegger2021machine}. This degrades the reward signals and destabilizes the learning. The problem is compounded in $3$D atomistic environments, where high-reward actions lie in a multimodal and discontinuous space. Here, small perturbations rarely improve the objective, but often disrupt chemically valid structures, rendering local exploration ineffective~\citep{rose2021reinforcement}. 

A true paradigm shift in molecular discovery requires the ability to explore the full chemical space from first principles—without relying on hand-crafted rules, curated datasets, or preconceived feature biases. Such an approach would learn viable chemistry entirely through exploration of chemical motifs leading to possible discovery beyond current human knowledge and intuition. As a key step toward this grand vision, we have been successful in training \textit{composition-generalizable} \ac{rl} agents capable of discovering stable 3D molecules across diverse chemical formulas.

Inspired by the \textsc{MolGym}~\citep{molgym1,molgym2} framework, we take a significant step forward toward training self-guided RL agents that can generalize across chemical space. Specifically, we introduce AtomComposer, a framework for isomer discovery in which an agent learns to generate stable 3D molecular structures given a pre-specified chemical composition, with no access to reference geometries during training.

\begin{figure}[t]
    \centering
\includegraphics[width=\textwidth]{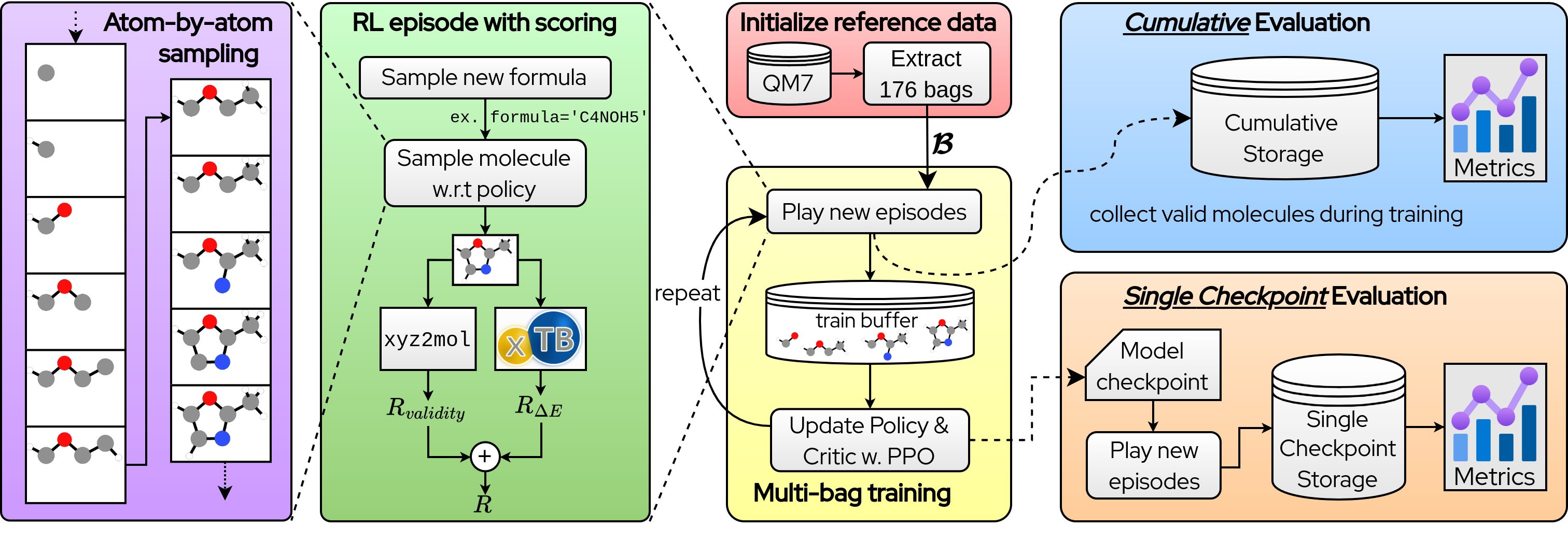}
    \caption{\textbf{The AtomComposer multi-composition training and evaluation workflow.} AtomComposer constructs isomer generation tasks by extracting chemical formulas from a reference dataset and introduces new terminal rewards based on validity and total energy. We evaluate the \ac{rl} agents' isomer discovery capabilities at just a \textit{single} checkpoint, as well as \textit{cumulatively} across the entire discovery campaign. \textbf{Crucially, no 3D structures are shown to the agent during training.}}
    \label{fig:rl_framework}
\end{figure}

Our success originated from a novel multi-composition training scheme and new reward schemes. We demonstrate that \ac{rl} can be effectively applied to isomer discovery, without overfitting to a fixed set of atoms as in prior work~\citep{molgym1,molgym2}. A visual abstract of the AtomComposer is provided in \cref{fig:rl_framework}. This resolves long-standing limitations and stagnation in \ac{rl} for \textit{tabula rasa} 3D atomic structure discovery and we summarize our main contributions as follows:
\begin{itemize}
    \item We introduce new terminal rewards based on energy and chemical validity, thereby training the agent to build stable and \textit{valid} molecules.
    \item  We propose a groundbreaking multi-composition training setup based on chemical compositions drawn from a broad chemical space derived from the \textsc{QM7} reference dataset, facilitating generalization across stoichiometries.
    \item We design a broader multi-bag evaluation scheme to facilitate benchmarking of online isomer discovery agents and assess various combinations of the proposed reward terms.
\end{itemize}

\section{Results}

\subsection{RL environment: Isomer search}
We trained the AtomComposer \ac{rl} agent to build stable and valid \textit{isomers} (i.e. different molecules with the same \textit{pre-specified} chemical formula) autoregressively (atom-by-atom), using a linear combination of reward terms based on quantum chemical energy evaluations and cheminformatics validity checks. The training framework is illustrated in \cref{fig:rl_framework}, along with the two separate evaluation schemes. 


\paragraph{Multi-composition training}
We leveraged the QM7 dataset~\citep{qm7-blum, qm7.rupp} as a \textit{reference dataset}, and used it solely to derive a bag set\footnote{Notice that "formula" and "bag" are used interchangeably throughout this article, as they carry the same physical meaning. So, to clarify, a \textit{bag set}, $\boldsymbol{\mathcal{B}}$, is simply a collection of chemical formulas. Similarly, to describe how our agent generalizes across chemical formulas, we use words such as "multi-bag", "multi-composition", "stoichiometry-agnostic", etc., depending on the context.}, $\boldsymbol{\mathcal{B}}$, of chemically valid molecular compositions for multi-composition training; the agent is never exposed to any reference geometries or structural information. In this role, QM7 provides an exhaustive enumeration of chemically feasible molecular formulas up to seven heavy atoms for the chosen element types, defining a well-controlled domain for studying composition-generalizable, purely online reinforcement learning in 3D. We find that this set already provides sufficient compositional diversity to learn the desired multi-bag behavior. Since all such formulas are enumerated at this scale, explicitly generating compositions on-the-fly, such as the stochastic bag sampling employed in \cite{molgym2}, is unnecessary. Moreover, restricting to this regime enables reliable evaluation against known isomer spaces without the ambiguity and computational overhead that would arise for substantially larger datasets such as QM9~\citep{doi:10.1021/ci300415d, ramakrishnan2014quantum}.

In practice, training rollouts are performed synchronously by a collection of $N_w$ workers, each endowed with a uniquely randomized iterable of the bag set $\boldsymbol{\mathcal{B}}_w=\text{permutation}_w(\boldsymbol{\mathcal{B}})$. When worker $w$ has generated a molecule for a particular bag (or failed to do so), it simply proceeds to the next bag in its bag set.

\paragraph{Autoregressive molecule sampling} The molecule construction process is framed as a sequential decision-making task, where, after sampling an initial bag of atoms $\mathcal{B}_0$, an agent iteratively selects and places atoms in $3$D space to incrementally build the molecule.
In RL terms, the agent observes the state $s_t=(\mathcal{C}_t, \mathcal{B}_t)$ consisting of the current molecular canvas $\mathcal{C}_t$ (i.e. the molecule built so far) and the remaining atom bag $\mathcal{B}_t$. The agent's action $a_t=(e_t, x_t)$ involves choosing an atom $e_t \in \mathcal{B}_t$ and assigning its $3$D position $x_t \in \mathbb{R}^3$ leading to the deterministic transition to the next state $s_{t+1} = (\mathcal{C}_{t+1}, \mathcal{B}_{t+1})$, where
$$\mathcal{C}_{t+1} = \mathcal{C}_t \cup \{(e_t, x_t)\}, \quad B_{t+1} = B_t \setminus \{e_t\}.$$
This process continues until the bag is empty and a complete molecule $\mathcal{C}_T$ has been formed. 
The distribution over molecules constructed in this autoregressive process is given by
\begin{equation}\label{eq:factorization}
    p(\mathcal{C}_T | \mathcal{B}_0) = \prod_{t=0}^{T-1} \pi_{\theta}(a_t | s_t),
\end{equation}
where $\pi_{\theta}(a_t | s_t)$ is the agent's probabilistic policy governing the placement of atom $e_t$ at position $x_t$, given the current molecular state $s_t$. This formulation captures the conditional nature of molecule construction starting from the initial bag $\mathcal{B}_0$.

Notably, in this environment, the agent must implicitly learn to construct valid molecules, as no explicit validity constraints are imposed during generation. Also, atoms are sampled without replacement, and their positions remain fixed after placement. The randomness in the generation process comes solely from the agent's policy, as the environment transitions are fully deterministic. As such, the molecule-building task can be formulated as a fully observable, finite-horizon Markov Decision Process (MDP) with a hybrid discrete-continuous action space, where the episode length is determined by bag size.

\paragraph{Reinforcement Learning objective}
The agent's stochastic policy $\pi_{\theta}(a_t | s_t)$ is optimized in search of the optimal parameters $\theta$ that maximize the expected discounted sum of future rewards (known as \textit{return}) from any given state,
\begin{equation}
    V^{\pi}(s_t) = \mathbb{E}_{\pi_{\theta}}\left[\sum_{t'=t}^T \gamma^{t'} r(s_{t'}, a_{t'})\right],
\end{equation}
where $\gamma \in (0, 1]$ is the discount factor and $r(s_t, a_t)$ is the reward received at time step $t$ for taking action $a_t$ in state $s_t$. So starting with an empty canvas at $t=0$, the agent must learn to maximize $J(\theta)=\mathbb{E}_{s_0 \sim \mu_0}[V^{\pi}(s_0)]$ with $\mu_0$ denoting the distribution over bags.

\paragraph{A new terminal reward structure}\label{Sec:Reward_terms}

In RL, reward design is often the single most critical factor determining success or failure. Whereas the original \textsc{MolGym} frameworks use \textit{per-step} rewards as shown in Box \ref{box:2.1b}b, we train agents which only receive reward at the terminal state, i.e. when the molecule is completed (Box \ref{box:2.1a}a). These rewards are determined based on quantum mechanical energies computed with the semiempirical method \verb|GFN2-xTB| \citep{bannwarth2019gfn2} and chemical valency checks via \verb|xyz2mol| \citep{kim2015universal}. \verb|GFN2-xTB| provides a favorable balance between computational efficiency and physically grounded energy estimates, enabling the large number of energy evaluations required for online reinforcement learning. Policy-gradient reinforcement learning does not require perfectly calibrated reward magnitudes; rather, the reward signal primarily needs to correlate with the underlying objective so that lower-energy configurations receive higher rewards on average. Previous studies have shown that \verb|GFN2-xTB| does preserve relative energy rankings for organic molecules in many cases \citep{seumer2023computational}, making it well suited to guide exploration of molecular configuration space within the QM7-derived organic domain considered here. Furthermore, the use of a physics-based energy evaluation avoids additional modeling assumptions introduced by learned neural surrogate models. Note that all molecular and atomic reference energies entering the atomization reward are computed consistently within the same \verb|GFN2-xTB| framework. In practice, the calculations correspond to the default electronic states of the respective species within the method (closed-shell singlet molecules and the ground-state configurations of isolated atoms). Since the reinforcement learning objective relies primarily on consistent relative energy ranking rather than absolute thermochemical accuracy, this choice provides a stable and internally consistent reward signal for the organic systems considered here.

The choice of terminal rewards stems from the fact that the temporal structure of our RL episodes is an artificial construct, introduced solely to enable the factorization of the agent’s molecular sampling policy (\cref{eq:factorization}), and the intermediate molecular states $\{\mathcal{C}_t\}_{t<T}$ visited during the episode are not necessarily chemically or energetically meaningful. To prevent training from being obscured by misleading or noisy signals from these partial, often non-physical intermediates, we introduce new terminal rewards that are only queried once the molecule is fully constructed, as shown in Box \ref{box:2.1a}a.

The validity reward presented here was introduced to stabilize learning under the stochastic construction policy. Because atomic coordinates are sampled with exploration noise and without prior knowledge of optimal geometries, intermediate distortions can lead to artificially high energies even for otherwise chemically sensible bonding patterns. The validity term therefore provides a geometry-agnostic structural signal that encourages chemically consistent connectivity before energetic refinement. This operational definition of validity reflects the valence conventions encoded in xyz2mol/RDKit and is therefore most appropriate for the organic chemistry regime considered here. It should not be interpreted as a universal chemical validity criterion for systems with non-standard bonding patterns (e.g., many organometallic or inorganic complexes). Importantly, this limitation arises from the specific validity signal used in the AV/AFV reward formulations rather than from the underlying reinforcement learning framework itself.

\begin{tcolorbox}[
    colback=BlueGreen!10, 
    arc=3mm, 
    auto outer arc,
    boxrule=0pt, 
    title={Box 2.1a: Terminal Rewards (ours)}, 
    center title,
    fonttitle=\bfseries,
    breakable
]\label{box:2.1a}
In this work, we introduce the following two terminal rewards:
\begin{itemize}
    \item \textbf{Atomization energy ($\mathcal{A}$)}: This reward is based on the negative difference between the potential energy of the final molecule $\mathcal{C}_T$ and the sum of potential energies of each of its constituent atoms in isolation:
    \begin{align}
        \Delta E = \left( \sum_{t=1}^T  E(e_t)\right) - E(\mathcal{C}_T), \ \ \  r_A(s_T)=
            \begin{cases}
                \Delta E + \frac{1}{2}(\Delta E)^2  & \text{if} \ \  \Delta E >0, \\
                \Delta E  & \text{if} \ \ \Delta E < 0,
            \end{cases}
    \end{align}
    where $\Delta E$ is the binding strength (positive is better) and the polynomial transformation provides extra resolution around high scoring molecules, as this allows the agent to differentiate between "good" and "really good" molecules.

    \item \textbf{Validity ($\mathcal{V}$)}: A boolean validity check based on whether the generated 3D structure can be successfully converted into a chemically consistent molecular graph using the \verb|xyz2mol| procedure, which infers bond connectivity and bond orders from Cartesian coordinates and returns an \verb|rdkit| \verb|Mol| object \citep{rdkit}:
    \begin{equation}
        r_V(s_T)=
        \begin{cases}
            1 \ \text{if } \mathcal{C}_T \ \text{is a valid molecule}, \\
            0 \ \text{else}.
        \end{cases}
    \end{equation}
    A structure is counted as valid if (i) the xyz2mol conversion succeeds, (ii) the resulting molecule forms a single connected component (i.e., is not fragmented), and (iii) no atom carries a formal charge.
    
\end{itemize}

\end{tcolorbox} 

\begin{tcolorbox}[
    colback=Periwinkle!15, 
    arc=3mm, 
    auto outer arc,
    boxrule=0pt, 
    title={Box 2.1b: Baseline Reward}, 
    center title,
    fonttitle=\bfseries,
    breakable
]\label{box:2.1b}
For comparison, the \textsc{MolGym} baseline used the following reward:
\begin{itemize}
    \item \textit{\textbf{Per-step Formation energy ($\mathcal{F}$)}}: In contrast to the terminal reward, reward can be assigned at every step throughout the episode and is given by the negative difference in energy between the resulting molecule $\mathcal{C}_{t+1}$ and the sum of energies of the previous molecule $\mathcal{C}_{t}$ and a new atom of element $e_t$
    \begin{equation}
        r_F(s_t, a_t) = \left( E(\mathcal{C}_{t}) + E(e_t) \right) - E(\mathcal{C}_{t+1}), \quad  t=0,...,T-1.
    \end{equation}
\end{itemize}

\end{tcolorbox}

\paragraph{Penalization of unrealistic molecules}
Whenever an atom is positioned unphysically close to any existing canvas atom, the episode is terminated prematurely and the agent receives a fixed failure penalty $R_{\text{kill}}$. In the current implementation, termination is triggered when any interatomic distance falls below a fixed, element-independent threshold. This deliberately simple cutoff provides a robust safeguard against catastrophic overlaps while avoiding additional element-specific heuristics in the environment; nevertheless, a more chemically resolved alternative would define atom-type–dependent exclusion radii from van der Waals or covalent radii to better reflect steric differences across elements.

This use of early termination of unpromising rollouts combined with purely terminal rewards (atomization and validity) can cause an untrained agent to struggle to obtain any reward whatsoever. Following standard practice in sparse-reward reinforcement learning, we therefore include a small constant positive per-step shaping reward to incentivize reaching the end of the episode, at which point the energy- and validity-dependent terminal reward is evaluated, although in principle this shaping term is unnecessary when $R_{\text{kill}} < 0$ and the discount factor satisfies $\gamma < 1$.

\subsection{Experiments}\label{Sec:experiments}

\begin{figure}[h!]
    \centering
    \includegraphics[width=1\linewidth]{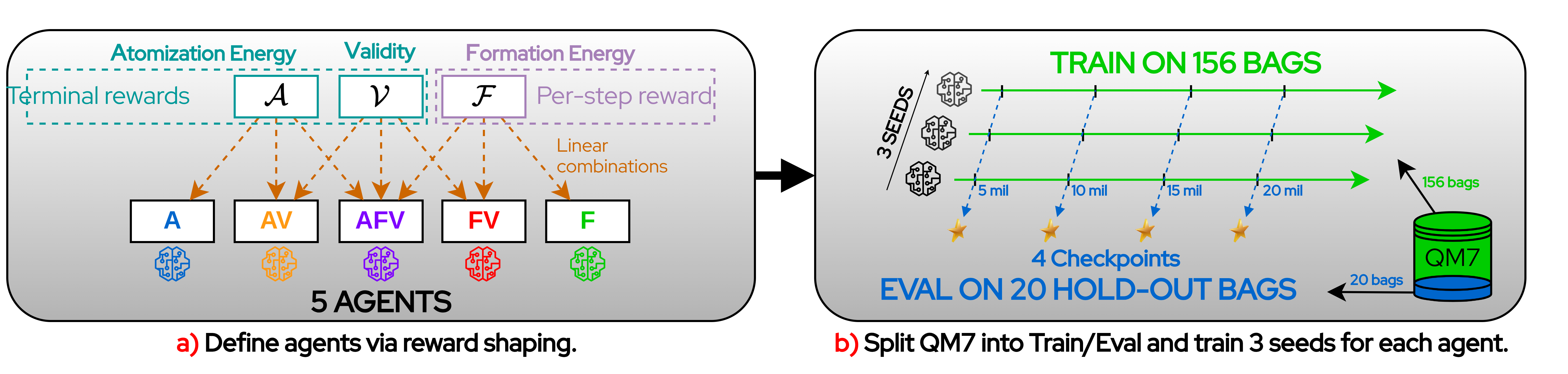}
    \caption{\textbf{Overview of experiments.} \textit{(a)} Agents are defined in terms of the reward functions with which they are trained. \textit{(b)} Training data comprises 156 QM7 bags, with 20 remaining bags held out for evaluation. During training ({\color{ForestGreen}green}) we save 4 checkpoints for each seed and perform out-of-sample evaluations ({\color{blue}blue}) for all checkpoints (5 agents, 3 seeds, 4 ckpts $\Rightarrow $ 60 evaluations in total).}
    \label{fig:exp_workflow}
\end{figure}

Figure \ref{fig:exp_workflow} illustrates our training and evaluation scheme. Through linear combinations of the 3 fundamental reward components ($\mathcal{A}$, $\mathcal{V}$, $\mathcal{F}$) introduced in Box \ref{box:2.1a}(a+b), we define 5 distinct reward functions {\textbf{\color{blue}A}}, \textbf{{\color{orange}AV}}, \textbf{{\color{ForestGreen}F}}, \textbf{{\color{red}FV}}, and \textbf{{\color{Plum}AFV}}, each corresponding to a separate \textit{agent} that is trained independently three times using different random seeds (the linear coefficients are shown in \cref{tbl:reward_coefs} in the Appendix). Generally, our analysis places particular emphasis on the agents {\textbf{\color{blue}A}} and \textbf{{\color{orange}AV}}, as these incorporate our newly proposed terminal rewards $\mathcal{A}$ and $\mathcal{V}$. Specifically, we address the following research questions:
\begin{itemize}
    \item \textbf{Q1: Comparison to previous work.} \textit{How do our agents ({\textbf{\color{blue} A}} \& \textbf{{\color{orange} AV}}) perform in comparison to previous work in online molecular discovery in $3$D?} This initial experiment evaluates their discovery capabilities in the single-bag generation paradigm where baselines are available.
    \item \textbf{Q2: Generalization ability.} \textit{Which reward functions generalize to our multi-bag setting and to out-of-sample (unseen chemical composition) generation in particular?} Here we broaden the evaluation scope relative to \textbf{Q1} by aggregating results across a random held-out split of bags represented in QM7 (see \cref{fig:exp_workflow}b), enabling a comprehensive comparison of reward signals at various stages of training. 
    \item \textbf{Q3: Exploration of chemical space.} \textit{Did our \textit{tabula rasa} agents rediscover the molecules from the QM7 reference dataset? Did they go beyond and even expand on this dataset?} Here we will interpret the training run as a discovery campaign and examine the complete pool of molecules obtained.
    \item \textbf{Q4: Property-directed finetuning.} \textit{Can the energy guided multi-bag agent be finetuned toward a specific molecular property — here the dipole moment — as a proof-of-concept for targeted molecular discovery?}
\end{itemize}

\begin{figure}[h]
    \centering
    \vspace{-2mm}
    \includegraphics[width=0.9\linewidth]{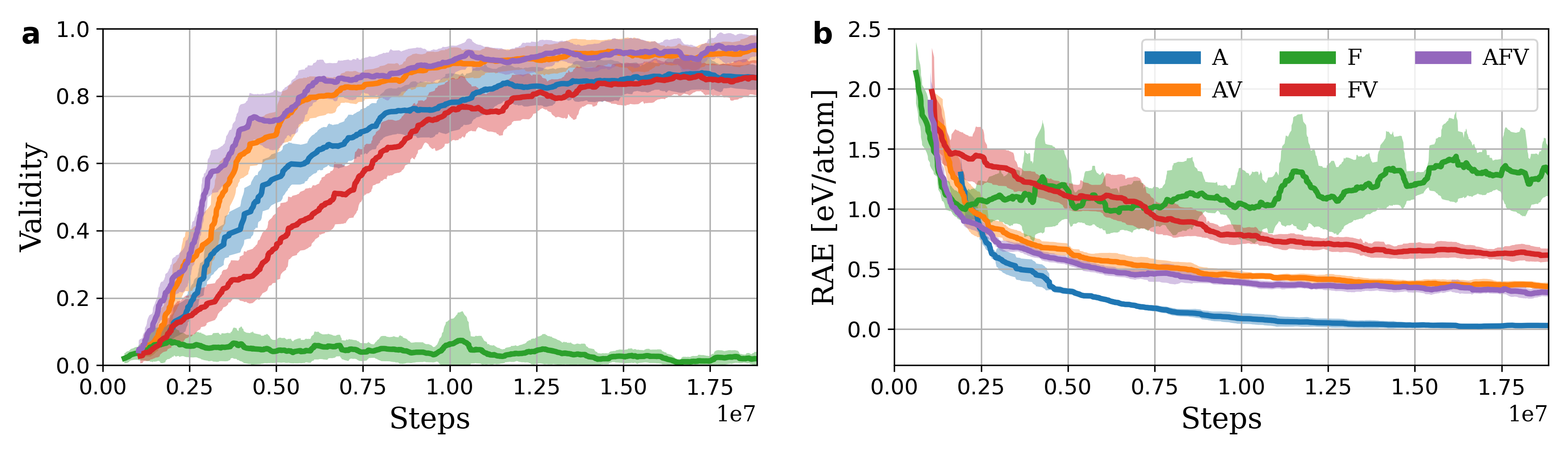}
    \vspace{-2mm}
    \caption{\textbf{Learning Curves (in-sample).} \textit{(a)} Validity and \textit{(b)} unrelaxed Relative Atomic Energy (RAE) of continuously collected {\color{ForestGreen}training rollouts}, plotted against total number of single-atom placements (environment steps). The RAE metric quantifies the excess energy relative to the average energies of QM7 molecules with the same chemical formula (see page \pageref{RAE_measure} for detailed metric definitions). Shading represents $\pm$1 standard deviation between the 3 seeds. \textbf{Notably, our newly introduced terminal reward terms, $\mathcal{A}$ and $\mathcal{V}$, enable significantly more stable training dynamics.}}
    \label{fig:learning_curves}
\end{figure}

While \cref{sec:Single-bag discovery,sec:Multi-bag aggregated evaluation,sec:property-directed-finetuning} (\textbf{Q1}+\textbf{Q2}+\textbf{Q4}) focus on evaluating single  checkpoints, \cref{sec:Cumulative discovery} (\textbf{Q3}) examines agent performance throughout the entire training process. Despite these differences in evaluation scope, all sections are based on the same training runs visualized in \cref{fig:learning_curves}.
Notice that agent \textbf{\color{blue}A}, which is trained solely with an energy-based terminal reward, consistently generates chemically valid molecules, with approximately 85\% validity during training rollouts (and >90\% validity on the 20 out-of-sample evaluation formulas presented in \cref{sec:Multi-bag aggregated evaluation}, \cref{fig:Q2}). This demonstrates that Agent \textbf{\color{blue}A}’s discovery of lower-energy isomers is not driven by the systematic generation of chemically non-viable structures.

\subsection{Q1: Comparison againts previous work - \textit{Single-bag discovery task}}\label{sec:Single-bag discovery}

\begin{table}[h]
    \centering
    \caption{\textbf{Q1: Single-bag discovery.} Our \textbf{\color{orange}AV} agent outperforms previous work (numbers taken from \cite{molgym2}) by discovering an order of magnitude more valid isomers for evaluation bags \textit{beyond} its training set. In contrast, the baseline agents were trained explicitly on the presented bags.
    }
    \label{tbl:single-bag-discovery}
    \begin{tabular}{c|lcccc}
        \toprule 


        
        & \textit{Training type:}    & \multicolumn{2}{c}{Single-bag training (on eval bag)}  & \multicolumn{2}{c}{QM7 multibag}   \\
        \cmidrule(r){3-4} \cmidrule(r){5-6}
        
        & \textit{Collection type:}  & \multicolumn{2}{c}{Cumulative argmax $\times$ 10 seeds}  & \multicolumn{2}{c}{Single CP stochastic}   \\

         \cmidrule(r){3-4} \cmidrule(r){5-6}

        &  \textit{Bag: }  \textbackslash \ \ \textit{Agent:} & \textsc{internal} & \textsc{covariant} & \textbf{{\color{blue} A} (ours)} &   \textbf{{\color{orange} AV} (ours)}  \\ 

        \midrule
         {\color{ForestGreen}\textsc{In QM7}}  &  $\text{C}_3\text{H}_8\text{O}$ & $4^{\dagger}$ & $\mathbf{8}^{\dagger}$  & \ \ \  \uncertainty{3.0}{0.0}    &  \  \uncertainty{3.0}{0.0} \\ 
        {\color{ForestGreen}\textsc{(train)}}   &  $\text{C}_4\text{H}_7\text{N}$ & $18$ & $25$  &  \ \ \uncertainty{13.0}{0.8}  \     &  \textbf{\uncertainty{36.7}{1.3}} \  \\ 
        \midrule
         {\color{Plum}\textsc{beyond}}   &  $\text{C}_3\text{H}_5\text{N}\text{O}_3$ & $35$ & $65$ & \ \  \uncertainty{49.0}{7.5}  \     & \ \textbf{\uncertainty{544}{46}}  \  \  \\ 

         {\color{Plum}\textsc{QM7}}  &  $\text{C}_7\text{H}_{10}\text{O}_2$ & $21$ & $85$   & \uncertainty{198.0}{24.9}       & \ \textbf{\uncertainty{808}{122}}  \\ 
        {\color{Plum}\textsc{dataset}}   &  $\text{C}_7\text{H}_8\text{N}_2\text{O}_2$ & $58$ & $118$   &\uncertainty{145.7}{39.7}     & \textbf{\uncertainty{1213}{212}} \  \\ 
        \bottomrule
    \end{tabular} \\
    \parbox{0.93\textwidth}{
    $^{\dagger} \text{C}_3\text{H}_8\text{O}$ is a small and fully saturated chemical formula and we only see 3 feasible positions for an oxygen atom on a 3-membered carbon chain: an $\text{OH}$ group on the first carbon atom, an $\text{OH}$ group on the central carbon atom, or an O between carbon atoms 1 and 2. Since both baseline agents reportedly discovered strictly more than 3 isomers without providing code for their uniqueness check, we suspect their numbers are mistakenly reported in all 5 cases, which only further emphasizes the improved discovery capabilities of our approach.}
\end{table}

We adopt the evaluation protocol from \cite{molgym2}, counting the number of valid constitutional isomers\footnote{Isomer counts are determined following the standard convention: unique SMILES strings are generated using \texttt{RDKit}~\citep{rdkit}, expressed in canonical form, and exclude isomeric information. Canonical SMILES is agnostic to the atom indexing. Tautomers are treated as distinct structures, whereas stereochemical variants are not differentiated. Thus, uniqueness is defined purely by the molecular connectivity graph. Note that InChI/InChIKey identifiers can be a complementary representation for database-level comparisons but were not used in this study.} discovered by our agents (\textbf{\color{blue}A} and \textbf{\color{orange}AV}) when deployed on a single bag.  \cref{tbl:single-bag-discovery} compares our results with those reported in prior work. Notably, although \textbf{\color{orange}AV} is not explicitly trained on certain formulas, it consistently discovers up to an order of magnitude more constitutional isomers than baseline agents on the last three formulas that exceed the scope of \textsc{QM7}, containing more than seven heavy atoms. In contrast, the \textbf{\color{blue}A} agent performs comparably to the baselines in terms of isomer count.

\begin{figure}[h!]
    \centering
    \includegraphics[width=1.0\linewidth]{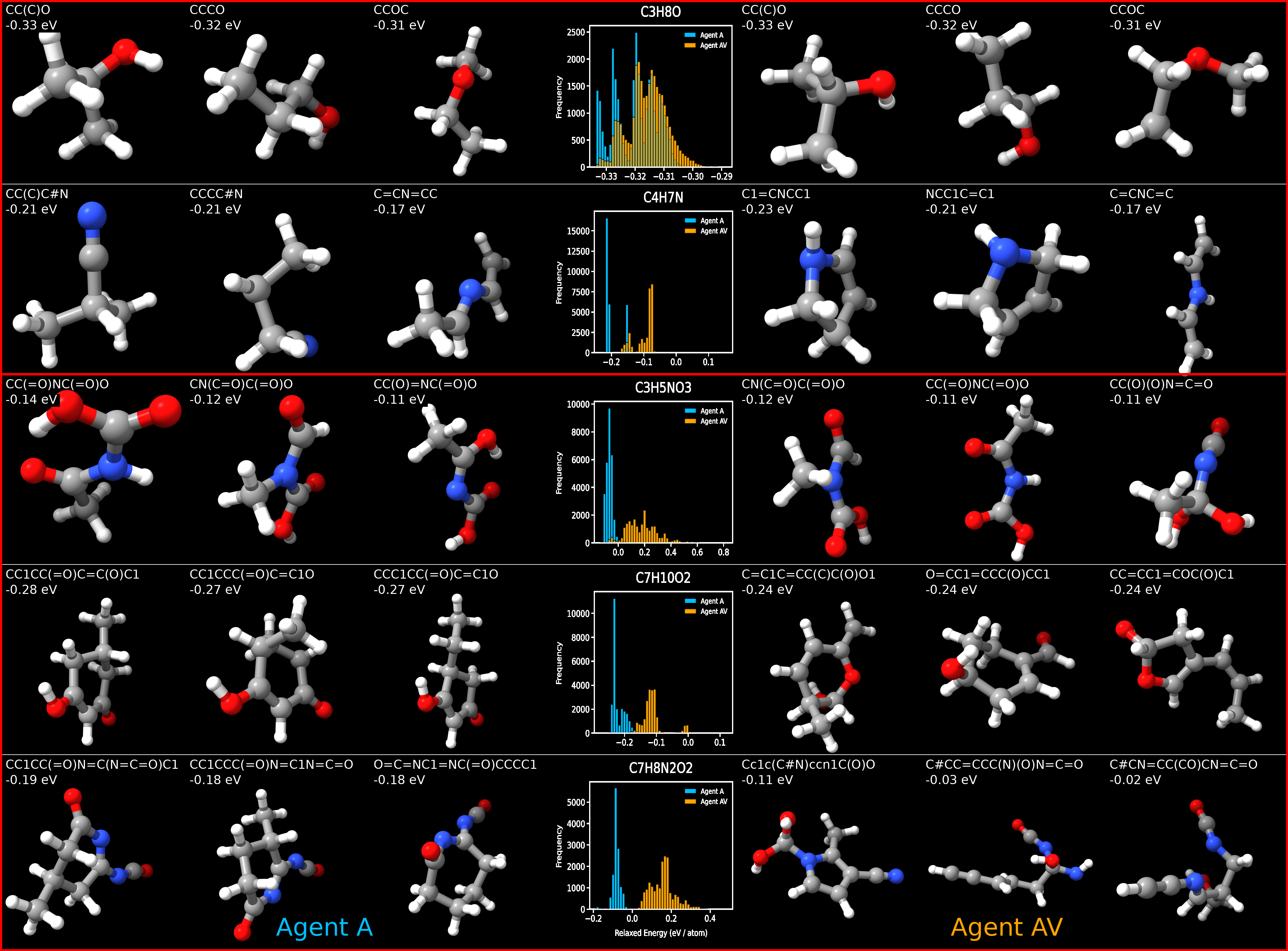}
    \caption{\textbf{Q1 visualizations.} Histograms of formation energy per atom (center column) together with top 3 best scoring molecules (lowest energy) after structural relaxation for Agent \textbf{\color{blue}A} on the left and Agent \textbf{\color{orange}AV} on the right. \textbf{Agent \textbf{\color{blue}A} samples molecules of significantly lower energies.}}
    \label{fig:Q1}
\end{figure}

In \cref{fig:Q1}, we visualize the high reward molecules, ranked by formation energy per atom after structural relaxation, and compare their energy distributions in the center column. While the \textbf{\color{orange}AV} excelled in breadth of discovery, the \textbf{\color{blue}A} agent - trained solely with energy-based rewards - tends to sample molecules with significantly better formation energies.

While the discovery statistics in \cref{tbl:single-bag-discovery} highlight the effectiveness of our training setup, reward formulation, and data collection strategy, it is important to note several key differences between our approach and the baseline methods:

\textbf{Baselines:} The \textbf{\textsc{internal}} and \textbf{\textsc{covariant}} agents from \textsc{MolGym} use a single-bag training paradigm. This is a costly approach that requires a separate training run for each conceivable molecular formula (bag). Additionally, the discovered isomer count is aggregated over 10 independent runs using different seeds. The molecules used for isomer counting are collected throughout the training (referred to here as \textit{cumulative} data collection), and the molecules are always generated by selecting the most likely action (i.e. $\arg \max_{\{a_t\}} \pi_{\theta}(a_t|s_t)$), resulting in just a single molecule at every checkpoint during training, thus relying solely on the gradual drift of the agent policy to achieve diverse sampling.

\textbf{Proposed scheme:} We adopt a multi-bag training strategy, using compositions derived from molecules in the \textsc{QM7} reference dataset, and evaluate discovery performance on the same test bags used by the baseline methods. Unlike the baselines however, we report results based on molecules sampled stochastically from the learned agent policy at a single checkpoint (CP). Concretely, we use the third checkpoint---taken after 15 million training steps---and sample 10,000 molecules per random seed for each test-time formula listed in \cref{tbl:single-bag-discovery} and \cref{fig:Q1}.

While RL policies may evolve over training, we find that the reported discovery count statistics are not highly sensitive to checkpoint selection. In particular, the multi-bag aggregated evaluation in \ref{sec:Multi-bag aggregated evaluation} (Q2) shows consistent performance across checkpoints for all successfully trained agents, indicating that the single-checkpoint results reported here are representative.



\subsection{Q2: Reward term comparison - \textit{Multi-bag aggregated evaluation}}\label{sec:Multi-bag aggregated evaluation}
While the previous experiment demonstrated the superior discovery capabilities of our agents, a much broader evaluation scheme is necessary for a robust comparison of reward signals. To achieve this, we carry out the experiment that was outlined in \cref{fig:exp_workflow}b ({\color{blue}blue}). Here, we aggregate results from a random split of 20 holdout formulas in the QM7 dataset, offering a more comprehensive assessment compared to single-bag evaluation.

\begin{figure}[h!]
    \centering
    \includegraphics[width=1.0\linewidth]{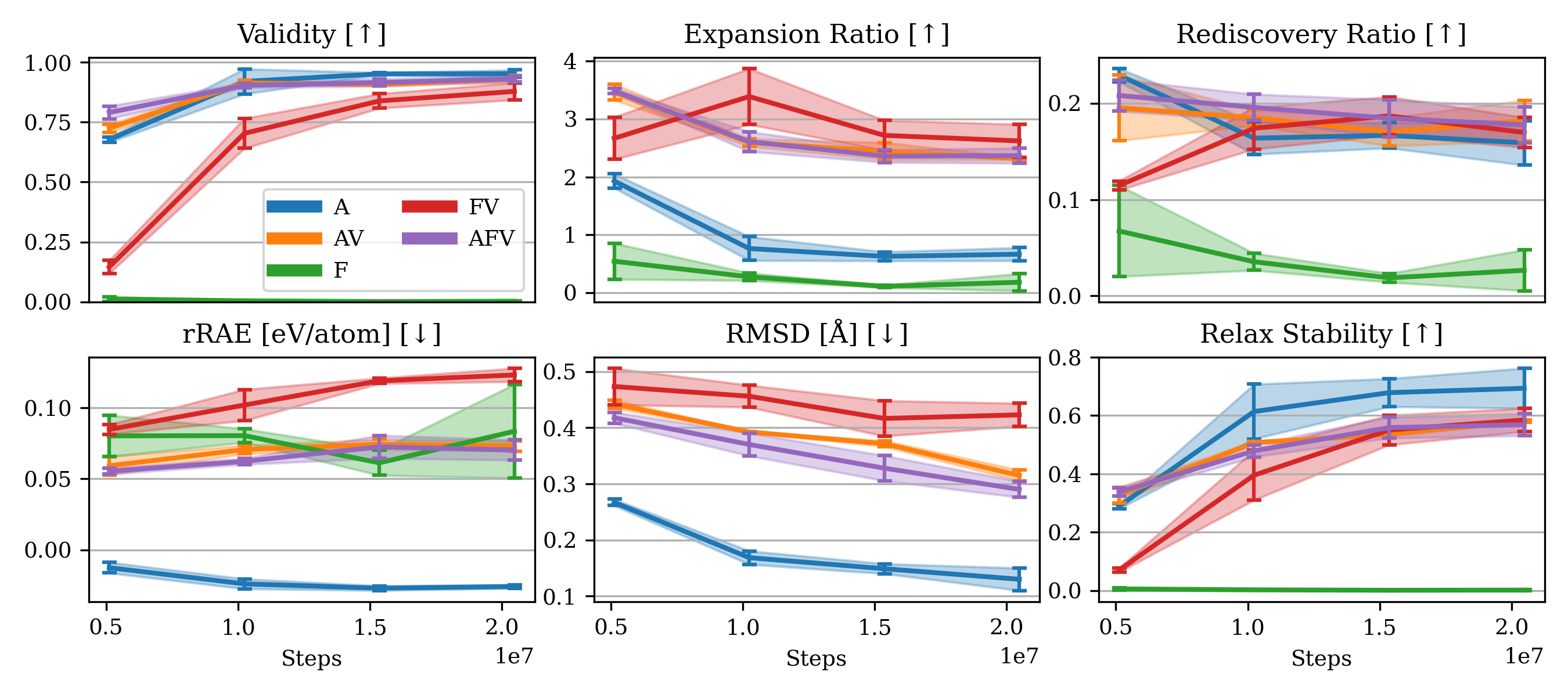}
    \caption{\textbf{Q2: Out-of-sample agent comparison.} We report discovery metrics (top row) and geometry metrics (bottom row) in the multi-bag evaluation setting outlined in \cref{fig:exp_workflow}b (see page \pageref{RAE_measure} for detailed metric definitions). Each point reflects a weighted average across 20 test bags. Error bars denote standard deviation across three random seeds. Results show that agent \textbf{\color{blue}A} consistently outperforms on 3D metrics, while \textbf{{\color{orange}AV}} and \textbf{{\color{Plum}AFV}} perform identically—highlighting the redundancy of $\mathcal{F}$ in our setup. Agent \textbf{\color{ForestGreen}F} fails to discover valid molecules due to excessive intra-episode rewards. Flat rediscovery and expansion metrics suggest no mode collapse, but agents fail to turn this into continued improvement toward more stable molecules.}

    \label{fig:Q2}
\end{figure}

For each test bag $\mathcal{B}_i$, $i=1,2,...,20$ we now sample $N_i = P \cdot N_i^{\text{ref}}$ molecules, where $N_i^{\text{ref}}$ is the number of isomers in the reference dataset for $\mathcal{B}_i$, and $P = 100$ is a proportionality factor. This scaling ensures that the number of sampled molecules reflects the expected isomer diversity for each bag. In practice this corresponds to several thousand sampled molecules per formula, which we found sufficient for stable discovery statistics while keeping evaluation computationally tractable. The results, including standard deviations across three seeds, are presented in \cref{fig:Q2}. Note that to obtain these metrics we first calculated these statistics for each bag individually and then aggregated across all hold-out bags using a weighted average according to $N_i$.

From \cref{fig:Q2} we make the following observations:
\begin{itemize}
    \item \textbf{>90\% validity} for agents that use either $\mathcal{A}$ or $\mathcal{V}$, with only \textbf{{\color{red}FV}} being slow to reach these levels. This level of validity is more than sufficient in an online reinforcement-learning context, where exploratory actions are essential for discovery and perfect validity would likely reflect overly conservative policies. Despite these high validities, we note that only $60\%-70\%$ of the generated molecules were "Relax Stable" (bottom right), meaning that relaxation did not alter the geometries significantly enough to change their bond connectivity and thus their SMILES representation.
    \item \textbf{Agent \textbf{\color{blue}A} dominates on $3$D metrics (same as Q1).} This confirms our observation from \textbf{Q1} that the terminal atomization energy signal, $\mathcal{A}$, best facilitates discovery of low energy structures. In fact, not only is it better than all other agents across all checkpoints, it is also better than the molecules present in the QM7 reference dataset \textit{on average}.    
    \item \textbf{Agent {\color{ForestGreen}F} does not learn to create valid molecules.} The per-step formation-energy reward provides strong local signals at each atomic placement, encouraging the agent to greedily stabilize intermediate structures without considering whether the resulting partial geometry can ultimately yield a chemically valid molecule once the bag is exhausted. Because valid molecules are only evaluated at the terminal state, these strong intra-episode rewards effectively bias the agent toward locally favorable but globally inconsistent construction strategies, preventing successful completion of full molecules.
    \item \textbf{{\color{orange}AV} and \textbf{{\color{Plum}AFV}} perform identically.} This shows that the per-step formation energy, $\mathcal{F}$, is a redundant signal in our setup and does not contribute significantly to justify the increase in energy queries within each episode.
    \item \textbf{Relaxed energy metrics don't improve (rRAE).} While the raw energies of the generated molecules do improve during training (not shown here), this effect is washed out upon relaxation as shown in the \textit{rRAE} plot. Despite the generated molecules becoming less noisy, as seen from the \textit{RMSD} and \textit{Relax Stability} curves, the agents' abilities to select low-energy isomers appears stagnant. And while it seems positive that the RMSD continues decreasing throughout training, this is merely a bi-product of the gradual narrowing of the width of the univariate Gaussians in their action policies from \cref{eq:molgym_policy} (recall that in order to facilitate exploration, the agents add Gaussian noise to the three spherical coordinates $(r, \alpha, \psi)$ as they sample the $3$D position of the new atom). This narrowing improves geometric precision but does not alter the underlying isomer connectivity and therefore does not change the relaxed energy ranking.
    \item \textbf{Rediscovery \& Expansion metrics are flat.} On the positive side, this means that the agents don't display severe mode collapse, which would cause these numbers to decrease during training. The sustained discovery is likely a consequence of the entropy regularization term (\cref{eq:entropy_term}) on the two discrete actions that ultimately determine the connectively and hence the SMILES object.
\end{itemize}

Furthermore, while Agent \textbf{\color{blue}A} excels at discovering lower-energy conformations, its main trade-off relative to \textbf{\color{orange}AV} is reduced structural diversity rather than chemical viability. The strong energy bias of agent \textbf{\color{blue}A} favors a narrower subset of low-energy isomers, whereas \textbf{\color{orange}AV} balances energy optimization with broader exploration.

\subsection{Q3: Chemical space exploration - \textit{Training as a discovery campaign}}\label{sec:Cumulative discovery}

In the previous experiments, we evaluated agent performance based on stochastic rollouts from a single checkpoint - a simple and general scheme widely used in generative modeling (e.g., supervised distribution learning) - but one that overlooks the evolving nature of an RL agent’s policy and introduces arbitrariness due to checkpoint selection. To fully leverage this behavioral drift over the course of training, we instead store every molecule generated in a cumulative storage buffer as illustrated in \cref{fig:rl_framework} (blue), which allows us to track the discovery process across time.

\begin{figure}[h]
    \centering
    \includegraphics[width=\textwidth]{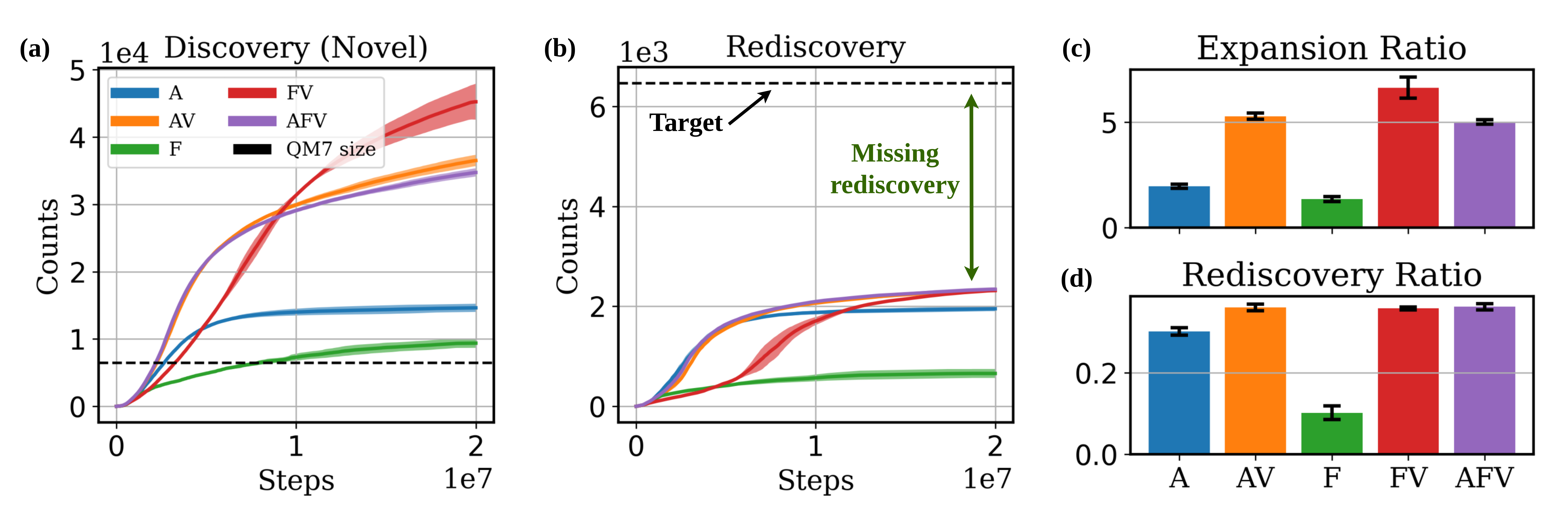}
    \caption{\textbf{Q3. Cumulative discovery campaign.} (a) Number of \textit{novel} SMILES discovered during training. (b) Number of QM7 SMILES rediscovered. (c)-(d) Total expansion and rediscovery \textit{relative} to the size of QM7. Although the agents are able to discover many novel molecules and expand on the QM7 dataset by several multiples, their rediscovery ratios are remarkably consistently capped around $40\%$, thus indicating a subclass of molecular structures inaccessible to our RL agents.}
    \label{fig:Q3:cum_total}
\end{figure}
\begin{figure}[h]
    \centering
    \includegraphics[width=0.70\textwidth]{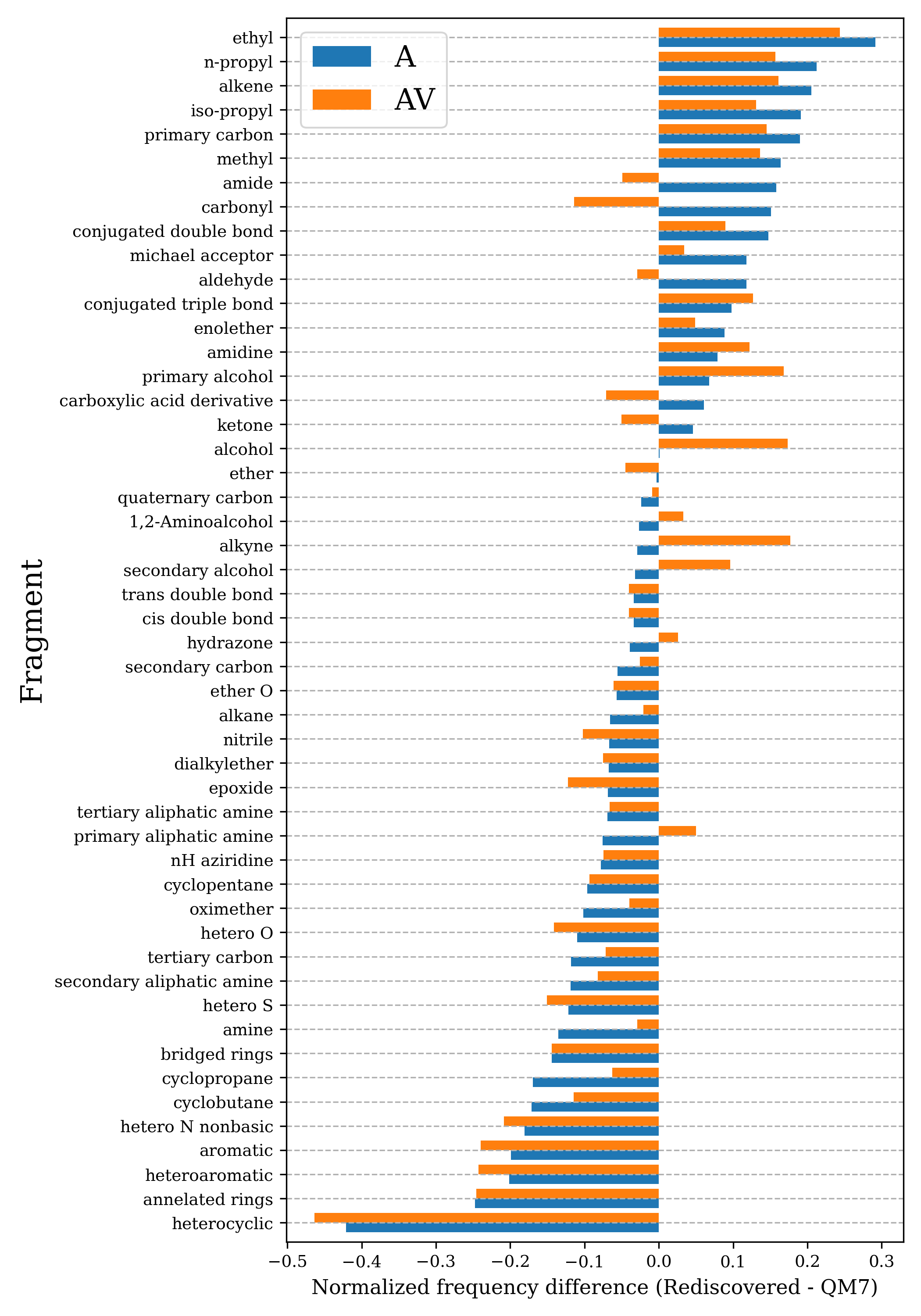}
    \caption{\textbf{Functional group analysis.} Normalized frequency difference of the 50 most common functional groups in the QM7 dataset between rediscovered molecules and the full QM7 reference set. To avoid rare functional groups dominating the extremes of the distribution, we report the normalized frequency difference $\tilde{\Delta}_f^i=(f_{\text{RL}}^i-f_{\text{QM7}}^i)/\sqrt{f_{\text{QM7}}^i}$ where $f_{\mathcal{D}}^i$ denotes the fraction of molecules in dataset $\mathcal{D}$ that contain at least one instance of functional group $i$. Each molecule contributes at most once per functional group, regardless of fragment multiplicity within the same molecule. Positive values indicate overrepresentation in the rediscovered set relative to QM7, while negative values indicate underrepresentation.}
    \label{fig:Q3:fragment}
\end{figure}

The cumulative discovery campaign is summarized in \cref{fig:Q3:cum_total}. During training, each agent discovers between 10,000 and 45,000 unique SMILES strings across the 156 training formulas (\cref{fig:Q3:cum_total}a). For comparison, the QM7 training subset contains only 6,465 molecules, so the number of generated molecules far exceeds the number of known reference structures. This relationship is captured by the expansion ratio shown in \cref{fig:Q3:cum_total}c, quantifying how many novel molecules the agent generates relative to the original QM7 set.

To assess the agent’s ability to reproduce known chemistry, we also count how many of the original QM7 molecules were rediscovered during training (\cref{fig:Q3:cum_total}b), with final rediscovery statistics shown in \cref{fig:Q3:cum_total}d. The rediscovery curves clearly plateau, indicating that further training does not yield additional rediscovered molecules. This plateau behavior across agents raises an important question:

\textit{Are there particular molecular substructures that our agents systematically fail to learn or explore?}

To investigate this, we analyze the rediscovery performance of our two most promising agents, \textbf{\color{blue}A} and \textbf{\color{orange}AV}, by comparing the sets of rediscovered molecules to the full QM7 training subset. We first pool rediscoveries across all three seeds to ensure robust statistics. Then, using the \texttt{exmol} package \citep{gandhi2022explaining}, we extract the most frequent functional groups in QM7 and compare their occurrence in rediscovered molecules vs. the full dataset. The results are shown in \cref{fig:Q3:fragment} where several key trends emerge. Among the most strongly underrepresented functional groups, we find
\begin{itemize}
    \item \textbf{Heterocycles and aromatic systems} (e.g., \textit{heterocyclic}, \textit{heteroaromatic}, \textit{aromatic}, \textit{annelated rings}): These involve complex ring topologies and delocalized bonding, which are difficult to construct via sequential atom placement.
    
    \item \textbf{Strained and fused rings} (e.g., \textit{cyclopropane}, \textit{cyclobutane}, \textit{cyclopentane}, \textit{bridged rings}, \textit{nH aziridine}): Geometrically strained or topologically complex rings are less favored due to their instability and the precise coordination required to assemble them.
    
    \item \textbf{Heteroatoms and functionalized amines} (e.g., \textit{hetero N nonbasic}, \textit{hetero S}, \textit{hetero O}, \textit{amine}, \textit{primary/secondary aliphatic amine}, \textit{oximether}): These groups introduce electronic and geometric diversity, making them harder to learn and reproduce, especially when their placement significantly affects molecular stability.
\end{itemize}

In contrast, several functional groups are overrepresented, indicating that the agents preferentially discover molecules with these features:

\begin{itemize}
    \item \textbf{Simple alkyl groups} (e.g., \textit{methyl}, \textit{ethyl}, \textit{n-propyl}, \textit{iso-propyl}, \textit{primary carbon}): These groups are structurally simple and frequently encountered in organic molecules, making them easy for the agent to generate and overrepresented in the rediscovered set.
    
    \item \textbf{Carbonyl-containing groups} (e.g., \textit{carbonyl}, \textit{amide}): These planar, well-defined functional groups may be favored by the energy model used during training, and are commonly found in stable molecules.
    
    \item \textbf{Unsaturated hydrocarbons} (e.g., \textit{alkene}): These motifs are structurally simple and energetically favorable, leading to their frequent appearance in generated molecules.
\end{itemize}

Overall, these observations suggest that the agent’s discovery policy is biased toward constructing small, aliphatic, and less topologically complex fragments. Meanwhile, more exotic, strained, or electronically diverse motifs are significantly underexplored.


\begin{figure}[hb!]
    \centering
    \includegraphics[width=0.7\textwidth]{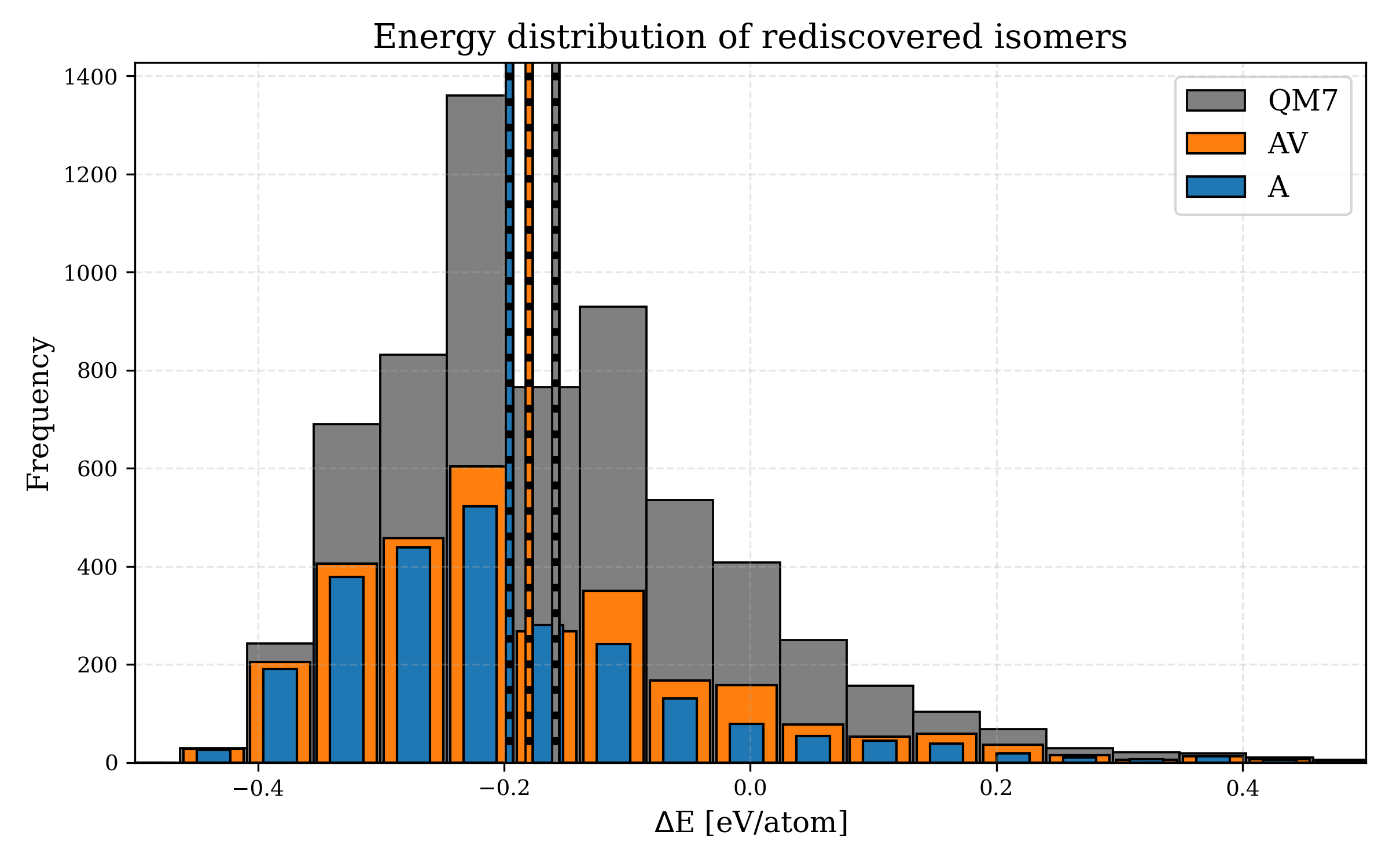}
    \caption{\textbf{Rediscovery energy distributions.} The figure shows the formation energy distribution of all QM7 training molecules (grey), together with the energy distribution of rediscovered molecules for the two agents \textbf{\color{blue}A} and \textbf{\color{orange}AV}, with mean values shown vertically. Despite rediscovering less than $50\%$, the RL rediscovered energies are actually better (more negative) than the QM7 average.}
    \label{fig:Q3:histogram}
\end{figure}
Finally, in \cref{fig:Q3:histogram}, we examine the distribution of formation energies for the rediscovered molecules. 
Despite recovering less than 50\% of the QM7 training set, both agents preferentially rediscover molecules with lower-than-average (i.e., more negative) formation energies. 
Notice again that this comparison is restricted to molecules within QM7, despite the agents having vastly expanded beyond it.

\subsection{Q4: Property-directed finetuning}\label{sec:property-directed-finetuning}
\begin{figure}[b!]
    \centering
    \includegraphics[width=\textwidth]{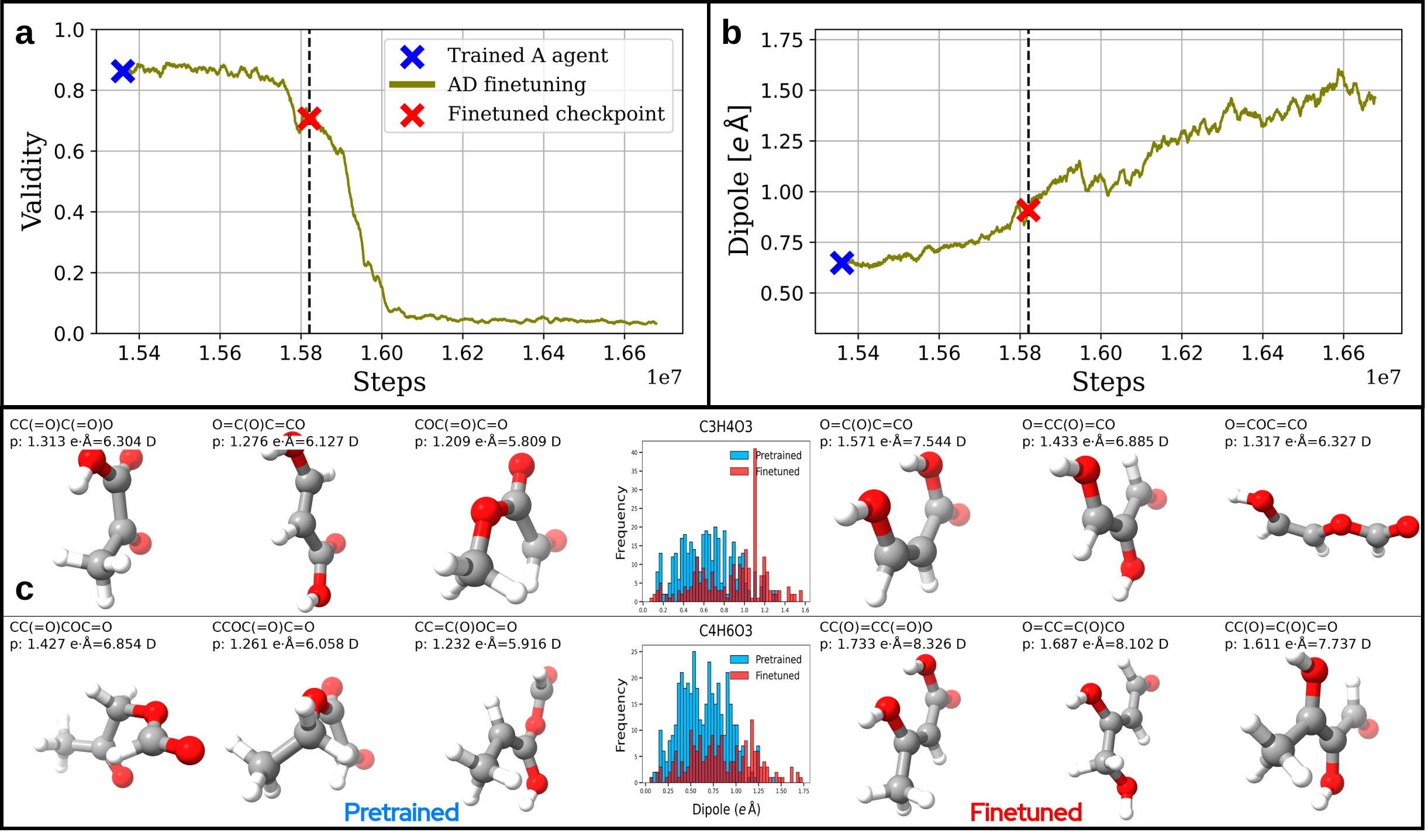}
    \caption{\textbf{Q4. Property-directed finetuning.} 
    (a) Validity and (b) dipole moment magnitude of training rollouts during finetuning, plotted against environment steps. A reward coefficient schedule gradually introduces the dipole moment reward alongside the existing atomization energy term. The red cross marks the evaluation checkpoint selected prior to significant validity collapse. 
    (c) Relaxed dipole moment distributions for the pre- and post-finetuning agents on the two formulas where finetuning improved performance (H4C3O3 and H6C4O3), together with the top 3 highest-dipole molecules discovered by each agent.}
    \label{fig:Q4-training}
\end{figure}

Liquid electrolytes are a critical component of lithium-ion batteries, where the ability of the solvent to dissolve lithium salts---governed in part by its dielectric constant---directly influences electrochemical performance. The molecular dipole moment is a key contributor to the dielectric response of a solvent, and carbonate-based solvents such as ethylene carbonate (EC) and propylene carbonate (PC) are favored in part due to their high dipole moments and correspondingly high dielectric constants. Discovering novel isomers of electrolyte-relevant molecular families with enhanced dipole moments could therefore be a useful proxy objective in the search for next-generation electrolyte solvents.


\paragraph{Training} To demonstrate that our framework can be directed toward such a practically relevant objective, we performed a finetuning experiment in which a pretrained \textbf{\color{blue}A} agent checkpoint (taken at 15 million environment steps, corresponding to the third checkpoint used in \textbf{Q1}) was further trained with an additional dipole moment reward term. We deliberately finetune the energy-only agent (\textbf{\color{blue}A}) rather than \textbf{\color{orange}AV}, as thermodynamic stability is a primary design criterion for battery electrolyte solvents — candidate molecules must remain chemically intact across wide electrochemical potential windows and over repeated charge-discharge cycles. This contrasts with applications such as drug discovery, where broader structural exploration may be prioritized over strict energetic stability, making the validity-augmented agent (AV) a more natural starting point. The dipole moment magnitude, computed via \verb|GFN2-xTB|, was incorporated as a scalar reward to be maximized alongside the existing atomization energy reward, requiring no modification to the underlying environment or policy architecture. To avoid destabilizing the critic, which must remain calibrated to predict total episode returns, we introduced a reward coefficient schedule that linearly ramps the dipole moment coefficient from 0 to 2 over 2,500 training iterations, corresponding to approximately 1.28 million environment steps. This is equivalent to the full x-axis in \cref{fig:Q4-training}(a+b) which illustrate the finetuning dynamics. The dipole moment of generated molecules increases steadily throughout finetuning, confirming that the agent successfully learns to optimize for this new objective. However, chemical validity degrades sharply as finetuning progresses, eventually collapsing almost entirely. We attribute this to the dipole moment reward dominating the atomization energy signal at later stages of finetuning; a more conservative reward coefficient cap would likely mitigate this issue. The post-finetuning evaluation checkpoint was therefore selected prior to significant validity collapse (red cross in the figure).


\paragraph{Evaluation} We evaluated the finetuned agent on a set of 10 carbonate and ether formulas representative of known battery electrolyte solvents, sampling 500 molecules per formula from both the pre- and post-finetuning checkpoints and performing structural relaxation on all generated structures. For 2 out of 10 formulas (H4C3O3 and H6C4O3, corresponding to ethylene carbonate and propylene carbonate), the finetuned agent discovered molecules with notably higher dipole moments than the baseline agent, as shown in \cref{fig:Q4-training}(c). Complete evaluation results for all 10 formulas are provided in Appendix~\ref{app:finetuning}.

These results serve as a proof-of-concept that the proposed RL framework can be directed toward technologically relevant molecular discovery objectives through reward finetuning alone. In a more complete materials discovery pipeline, the dipole moment reward could be complemented by additional objectives such as electrochemical stability window, viscosity proxies, or synthetic accessibility scores, enabling multi-objective optimization across competing electrolyte design criteria — all within the same terminal reward formulation and without modifying the underlying environment or policy structure. The observed validity degradation at later finetuning stages highlights the importance of careful reward coefficient scheduling in multi-objective settings. A thorough investigation of such trade-offs is beyond the scope of the present study; we refer the reader to the broader multi-objective reinforcement learning literature for best practices in balancing competing reward signals~\citep{hayes2022practical, dai2023safe}.

\section{Discussion}


We presented AtomComposer, an autoregressive, multi-composition reinforcement learning (RL) agent for $3$D isomer discovery, trained purely online across a large set of molecular formulas derived from the QM7 dataset. This represents a significant advancement over composition-specific RL agents for 3D structure discovery that have been developed in recent years. Our method enables, for the first time, the autonomous exploration of broad chemical spaces without reliance on curated datasets, paving the way for serendipitous molecular discoveries.

Beyond improved quantitative performance, the present work represents a conceptual shift from previous MOLGYM-style approaches: although we build upon the same actor–critic backbone, the learning objective and training regime are fundamentally restructured rather than merely reweighted or retuned. In particular, the shift from per-step formation-energy rewards to purely terminal reward evaluation fundamentally alters the credit-assignment structure of the task. In the original formulation, intermediate energy differences guide local atomic placements, encouraging myopic improvements that do not necessarily support coherent full-molecule assembly. In contrast, our terminal reward scheme requires the agent to plan across the entire episode horizon, as reward is obtained only once the complete molecular structure has been constructed. This transforms the learning problem from incremental local stabilization into a genuinely global structure-generation task.

Furthermore, the explicit integration of chemical validity into the terminal reward does not act as a post hoc filter, but reshapes the effective reward landscape during learning. While the energy-only agent (A) already produces a high fraction of valid molecules—since grossly unphysical structures tend to incur large energetic penalties—the validity term plays a distinct role. By providing a geometry-agnostic structural signal, it reduces the risk that moderate spatial noise obscures chemically meaningful configurations through overly harsh energy penalties. As a result, validity shaping supports broader exploration of structurally diverse isomers, rather than restricting the policy to a narrow subset of extremely low-energy configurations.
The observed failure of the purely per-step agent (F) in our multi-bag evaluation highlights that intermediate reward shaping does not merely introduce noise, but can actively impede the coordinated long-horizon planning required for valid molecule completion.

Our method used smaller learning rates, higher entropy coefficients, and a multi-bag training scheme that improved chemical and geometric diversity to achieve generalization unlike previous methods. Crucially, the introduction of multi-composition training changes the learning regime from composition-specific optimization to policy generalization across stoichiometries. Whereas single-bag agents learn a construction strategy tailored to one predefined formula, our agent must develop transferable placement heuristics that operate across diverse chemical compositions. These design choices prevented premature convergence to locally optimal policies and enabled broader exploration of molecular space. As a result, the agent learned to generate a diverse set of valid isomers—even for unseen formulas—and significantly outperformed single-bag agents.


Notably, we found that terminal rewards yielded more stable learning than per-step rewards, despite common assumptions favoring the latter for improved credit assignment. This likely stems from the chemical implausibility of intermediate structures in an autoregressive construction process, which renders stepwise energy signals noisy or misleading. However, a major limitation of terminal rewards—and of the current RL formulation more broadly—is the sparse and delayed nature of the learning signal, which leads to inefficient training and diminishing returns with extended optimization. These effects point to a fundamental scalability bottleneck in the present setup. 

This scalability bottleneck is not accidental, but rather reflects a deliberate tradeoff in the present study. Specifically, we restrict the agent to a fixed atom-bag formulation, such that each episode terminates only once all atoms in the bag have been placed. While this strictly atom-by-atom construction process limits flexibility, it enables a well-defined task with an exhaustively characterizable reference space, which is essential for quantitatively assessing discovery, rediscovery, and expansion in a tabula rasa reinforcement-learning setting.

A natural extension of the present AtomComposer formulation would be hybrid reward shaping, in which intermediate rewards are provided only when partial structures satisfy coarse chemical plausibility constraints (e.g., bond consistency or steric feasibility), thus yielding a potentially more sample efficient hybrid between AtomComposer (terminal rewards) and \textsc{MolGym} (per-step rewards). 
However, even chemically informed intermediate rewards do not address the fundamental constraint of the current AtomComposer setup; the strictly append-only and irreversible construction process. While reward shaping may guide local decisions, it does not permit correction of earlier suboptimal placements once committed. The irreversible nature of the current action space prevents the agent from revising or removing poorly placed atoms. As a result, partially constructed molecules that contain suboptimal early decisions cannot be recovered, forcing the agent to abandon potentially high-reward rollouts and restart from scratch. This limitation becomes increasingly restrictive for larger systems and helps explain both the observed plateau in rediscovery rates and the systematic under-representation of topologically complex motifs such as aromatic and heterocyclic rings.

These challenges are further compounded by a fundamental tradeoff between exploration and validity. While stochastic spatial noise is essential for promoting diversity and escaping local optima, energy-based reward signals are highly sensitive to geometric perturbations, such that excessive noise degrades both chemical validity and reward fidelity. From a computational perspective, longer episode horizons, sparser rewards, and increasingly costly quantum chemical evaluations collectively limit scalability to larger molecules and materials systems. Addressing these issues will require more flexible and expressive action spaces and representations, such as reversible or graph-level editing operations, fragment- or motif-based construction schemes, and policy parameterizations that reduce reliance on raw spatial noise. 
Complementary strategies, including mechanisms to discourage structural redundancy across rollouts or the incorporation of pretrained components that are subsequently refined through online interaction, could also substantially improve sample efficiency. Especially when extending the framework to broader element sets beyond QM7, a practical direction is to initialize atom-type representations using pretraining on less open-ended objectives, such as property prediction or offline RL constructed from existing molecular datasets, and subsequently refine them through online interaction. Such approaches would allow chemically meaningful atom embeddings to be acquired in a data-efficient manner, while still allowing the full 3D construction policy to be optimized purely through terminal reward signals. Together, these directions provide clear pathways toward broader and more efficient coverage of complex chemical spaces, while preserving the principled, first-principles nature of the proposed RL framework.

As discussed in the Methods \cref{sec:agent_policy} (\textit{Hierarchical action scheme}), the current policy samples the internal spherical placement coordinates using a factorized parameterization, which limits its ability to explicitly represent correlated geometric relationships between atomic positions. This design choice favors a minimal and scalable action space, but constrains the expressiveness of atom-by-atom placement decisions, particularly for topologically complex or highly coordinated motifs. Importantly, overcoming this limitation does not require imposing empirical chemical rules or bond statistics; rather, increasing policy expressiveness—through jointly or conditionally parameterized spatial actions, reversible placement operations, or higher-level construction primitives—would allow such correlations to be learned directly from terminal reward signals while preserving the generality and first-principles nature of AtomComposer as a scalable platform for autonomous chemical discovery.


A design choice in our formulation is that the agent explicitly places all atoms, including hydrogens, as it operates directly on 3D atomic point clouds without predefined bond orders. While many cheminformatics pipelines generate heavy-atom backbones first and subsequently add hydrogens using standard valence rules, such approaches rely on explicit bond type information and are therefore not directly compatible with our setting. 
Incorporating explicit bond information could enable post-hoc hydrogen placement and reduce the action-space complexity, and we leave this as an interesting direction for future work.

As demonstrated in \cref{sec:property-directed-finetuning} (\textbf{Q4}), the modular and terminal reward formulation adopted here is naturally extensible to property-directed molecular discovery. The dipole moment finetuning experiment shows that additional property-based objectives can be incorporated alongside stability without altering the underlying environment or policy architecture. More ambitious objectives---such as solubility, toxicity, or synthetic accessibility---could be incorporated in the same manner, supporting future exploration of trade-offs between competing molecular objectives.

\section{Methods}

\subsection{Agent policy}\label{sec:agent_policy}

\paragraph{Hierarchical action scheme}
We parametrize the agent's neural network policy following the \textit{internal} agent structure in previous work \cite{molgym1}, where at each decision step the agent breaks its action up into a hierarchical cascade of subactions, with each selected subaction being explicitly used as condition for the following. The three subactions are as follows: 
\begin{itemize}
    \item \textit{a)} selection of focal atom on the current canvas, around which a new atom will be placed,
    \item \textit{b)} selection of an element from the remaining bag,
    \item \textit{c)} $3$D placement of the new atom using a spherical coordinate system $(d, \psi, \alpha)$, where the coordinate axes are derived from distance vectors from the focal atom to - and among - its nearest neighbors.
\end{itemize}
In summary, the total agent policy is represented by the following factorization
\begin{equation}\label{eq:molgym_policy}
    \pi_{\theta} (d, \alpha, \psi, e, f| s) = p(d, \alpha, \psi | e, f, s)p(e|f, s)p(f|s),
\end{equation}
where the element $e$ and focal atom $f$ are sampled from categorical distributions and each of the coordinates $(\psi, \alpha, d)$ (which together define a unique mapping to a $3$D location for the new atom) are sampled from continuous distributions (univariate Gaussians).

While not immediately evident from the equations above, we must highlight the following potential conceptual flaws in the specific code implementation of this policy:
\begin{itemize}
    \item \textit{a)} The $3$D position distribution $p(d, \alpha, \psi | e, f, s)$ from \cref{eq:molgym_policy} actually imposes an assumption of independence between the internal spherical coordinates:
    \begin{equation}
    p(d, \alpha, \psi|e, f, s) = p(d|e, f, s) p(\alpha|e, f, s) p(\psi|e, f, s),
    \end{equation}
    since these variables are sampled independently rather than sequentially with mutual conditioning.
    \item \textit{b)} Even an infinitesimally small change in atomic positions can cause permutations in the nearest-neighbor ordering, thereby altering how prediction spherical coordinates $(\psi, \alpha, d)$ map to Cartesian canvas space.
\end{itemize}
In combination, these issues limit the agent's ability to make intentional and coordinated placement decisions with predictable outcomes. However, despite their potential significance, we proceed with this baseline implementation and leave the development of mitigation strategies for these policy limitations to future work.

\paragraph{Backbone message passing}
Instead of the \textit{invariant} backbone~\citep{schutt2017schnet} used in the baseline work \cite{molgym1}, we use its \textit{equivariant} counterpart ~\citep{schutt2021equivariant}, which additionally propagates vectorial features through its message-passing layers.
Although the spatial actions themselves only need rotational \textit{invariance}, given the nearest-neighbor-based internal coordinate system, invariant GNN architectures are known to have limited spatial and orientational awareness compared to equivariant architectures. This limitation has been underscored in subsequent work~\citep{molgym2} that demonstrated that equivariant architectures more effectively capture geometric relationships. Thus, enhancing the message-passing backbone by adopting the equivariant \textsc{PaiNN} architecture is well justified. Note, however, that within the neural network computation graph, the agent's spatial subactions $(d, \alpha, \psi)$ remain dependent exclusively on the invariant scalar tensor outputs rather than directly utilizing equivariant vectorial features.

\paragraph{Proximal Policy Optimization}
As shown in \cref{eq:molgym_policy}, action probabilities are explicitly modeled through a stochastic policy, making the framework amenable to the broader class of policy gradient methods \citep{williams1992simple, sutton1999policy}. These methods optimize the expected return by following the gradient of the policy’s parameters and are particularly effective when combined with a learned value function to form an Actor-Critic architecture \citep{konda1999actor}, which maintains both a stochastic policy (the actor) and a value function $V_{\theta}(s_t)$ (the critic). Specifically, we employ Proximal Policy Optimization (PPO) \citep{schulman2017proximal}, a widely used Actor-Critic algorithm designed to stabilize learning by preventing excessively large policy updates, as we shall see shortly.

At each iteration of PPO training, the agent first samples molecules according to its \textit{current} action policy, i.e. we record trajectory rollouts consisting of transition tuples $(s_t,a_t,r_t,s_{t+1}) \sim \pi_{\theta}$.
Based on this data buffer of freshly sampled rollouts, $\mathcal{B}_{\theta}$, PPO then enters a sequence of optimization steps, $k=1,2,3..,K$, where it performs gradient ascent on the following combined objective:
\begin{equation}
\mathcal{L}^{\textit{PPO}}(\theta)= \mathcal{L}^{\text{CL}}(\theta) - c_1 \mathcal{L}^{\text{V}}(\theta) + c_2 \mathcal{L^H}(\theta), \quad \quad c_1, c_2 \geq 0,
\end{equation}
where $\mathcal{L}^{\text{CL}}$ denotes the PPO \textit{clipped} surrogate objective, $\mathcal{L}^{\text{V}}$ is the value function loss, and $ \mathcal{L^H}$ represents the entropy regularization term.

The first objective term, $\mathcal{L}^{\text{CL}}(\theta)$, plays a central role in policy optimization, as it directly guides the policy toward selecting high-scoring actions. Rather than naïvely maximizing raw returns, PPO uses the Generalized Advantage Estimator (GAE) \citep{schulman2015high} to compute a smoothed signal of how advantageous an action was, relative to the expected value:
\begin{equation}
\hat{A}_t = \sum_{k=0}^{T-t-1} (\gamma \lambda)^k \delta_{t+k}, \quad\quad \text{with} \quad \delta_t = r_t + \gamma V_\theta(s_{t+1}) - V_\theta(s_t),
\end{equation}
where $\gamma \in [0, 1]$ is the discount factor and $\lambda \in [0, 1]$ is a smoothing parameter that controls the bias-variance trade-off. The resulting advantage term $\hat{A}_t$ captures the degree of \textit{positive surprise}, i.e. how much better (or worse) the observed return was compared to what the critic predicted.

After a few policy updates, the optimized policy $\pi_{\theta}$ can deviate significantly from the original policy that generated the data, which we from now on denote $\pi_{\theta_{\text{old}}}$. This mismatch can lead to unstable learning, as the objective is now being evaluated under a different distribution. To mitigate this, PPO introduces a clipped surrogate objective that limits how much the policy is allowed to change in a single update. The objective is defined as:
\begin{equation}
\mathcal{L}^{\text{CL}}(\theta)=\underset{(s_t, a_t) \sim \mathcal{B}_{\theta_{\text{old}}}}{\mathbb{E}}\left[\min\left(r^{\spadesuit}_t(\theta)\hat{A}_t ; \text{clip}(r^{\spadesuit}_t(\theta), 1-\epsilon, 1 + \epsilon) \hat{A}_t \right) \right], \quad  r^{\spadesuit}_t(\theta)=\frac{\pi_{\theta}(a_t|s_t)}{\pi_{\theta_{\text{old}}}(a_t|s_t)}.
\end{equation}
Here, the policy ratio $r^{\spadesuit}_t(\theta)$ quantifies how the probability of taking action $a_t$ under the updated policy compares to that under the old policy. The \verb|clip| operation ensures that this ratio stays within a trust region defined by $[1 - \epsilon, 1 + \epsilon]$, where $\epsilon$ is a small constant (commonly 0.2). This conservative update rule prevents excessively large policy shifts, thereby improving training stability. Intuitively, it balances learning progress - encouraging updates when $\hat{A}_t$ is large - with policy trustworthiness, by suppressing updates that would cause the policy to change too aggressively.

Next, the value loss term $\mathcal{L}^{\text{V}}$ is implemented as the mean squared error (MSE) between the predicted state-value estimates $V_{\theta}(s_t)$ and the empirical returns $R_t$. This term ensures that the critic network effectively predicts future returns, enabling more accurate advantage estimation in future iterations, while simultaneously distilling better representations into the shared backbone:
\begin{equation}
\mathcal{L}^{\text{V}}(\theta)=\underset{s_t \sim \mathcal{B}_{\theta_{\text{old}}}}{\mathbb{E}}\left[\left(V_{\theta}(s_t)-R_t\right)^2\right].
\end{equation}

Finally, the entropy term encourages exploration by penalizing overly certain action distributions: 
\begin{equation}\label{eq:entropy_term}
\mathcal{L^H}(\theta)=\mathcal{H}[\pi_{\theta}(e_t, f_t|s_t)] = \underset{(s_t, f_t, e_t) \sim \mathcal{B}_{\theta_{\text{old}}}}{\mathbb{E}} \left[- \log\pi_{\theta}(e_t, f_t|s_t) \right].
\end{equation}
Note that \textbf{entropy maximization is applied only to the two categorical subactions}: choosing the focal atom $f$ and selecting a new element $e$. The continuous spatial subactions $(d, \alpha, \psi)$, parameterized as univariate Gaussians, are exempt from entropy regularization to avoid inadvertently disrupting precise spatial predictions. Furthermore, to prevent premature convergence and encourage sustained exploration, we linearly increase the entropy coefficient, $c_2$, during training from 0.15 to 0.25.

\section*{Data and code availability}
The AtomComposer code, data, and instructions are available at \url{https://github.com/bhastrup/atomcomposer.}


\section*{Acknowledgements}
We acknowledge financial support from the Independent Research Fund Denmark with project DELIGHT, Grant No. 0217-00326B, and the Pioneer Center for Accelerating P2X Materials Discovery (CAPeX), DNRF grant number P3. We also acknowledge support from the Novo Nordisk Foundation Data Science Research Infrastructure 2022 Grant: A high-performance computing infrastructure for data-driven research on sustainable energy materials, Grant No. NNF22OC0078009. Finally, we thank Jan Jensen and Magnus Strandgaard for useful discussions on molecular scoring using \verb|xTB-GFN2| and \verb|xyz2mol|.

\bibliographystyle{unsrtnat}
\bibliography{references}

\clearpage
\appendix

\section{Training details}

\subsection{Hyper parameters}
In \cref{tbl:hyper-params} we show the hyper parameters used for model training. Additionally, we \colorbox{orange!20}{highlight} the individual hyper parameters whose values most crucially facilitated the training of a \textit{generalizable} agent (multi-component agent), as well as those \colorbox{ForestGreen!10}{unique to our setup}. In addition to our new reward structures and training schemes, the most noticeable difference from previous work is the use of larger data collections, smaller learning rates and higher exploration factors.
\setlength{\intextsep}{0pt} 

\begin{table}[ht]
    \centering
    \caption{Hyperparameters. {\color{blue} The neural agent policy consists of 242,351 parameters in total.}}
    \label{tbl:hyper-params}
    \begin{tabular}{c|lrl}
        \toprule 
       Category  &    Hyperparameter  &  Value & Code variable name \\
        \midrule
        &    Range $\left[ d_{\text{min}}, d_{\text{max}} \right]$ (Å) & [0.8, 1.8] & [\verb|min,max|]\verb|_mean_distance| \\
        \textsc{Rollout}  &    Workers & 8 & \verb|num_envs| \\
\rowcolor{orange!20}
        &    Env steps per PPO batch & 512 & \verb|num_steps_per_iter| \\
        & Fail reward & -3 & \verb|min_reward| \\
        \midrule
        &    Discount factor $\gamma$ & 1.0 & \verb|discount| \\
        &    GAE parameter $\lambda$ & 0.97 & \verb|lam| \\
        &    Value coefficient & 0.5 & \verb|vf_coef| \\
        &    Advantage clipping $\epsilon$ & 0.2 & \verb|clip_ratio| \\
       \textsc{Optimization}  &    Gradient Clipping & 0.5 & \verb|gradient_clip| \\
\rowcolor{orange!20}
        &    Learning rate (PPO-Adam) & 5e-5 & \verb|learning_rate| \\
        &    Minibatch size & 256 & \verb|mini_batch_size| \\
\rowcolor{orange!20}
          &    Entropy coef iter start & 0 & \verb|start_entropy_iter|  \\
\rowcolor{orange!20}
          &    Entropy coef value start & 0.15 & \verb|start_entropy| \\
\rowcolor{ForestGreen!20}
          &    Entropy coef iter end & 30,000 & \verb|final_entropy_iter| \\
\rowcolor{ForestGreen!20}
          &    Entropy coef value end & 0.25 & \verb|final_entropy| \\
        \midrule
        &   Layers &  3 & \verb|num_interactions| \\
       \textsc{PaiNN embedding}   &    Network width &  128 & \verb|network_width| \\
      & $r_{\text{cutoff}}$ (Å) & 5.0 & \verb|cutoff| \\
        \midrule
\rowcolor{ForestGreen!20}
       \textsc{QM7 splits}   &   Num training bags &  156 & \verb|..| \\
\rowcolor{ForestGreen!20}
        &    Num eval bags &  20 & \verb|n_test| \\

      \midrule
    \rowcolor{ForestGreen!20}                            & Dipole coef iter start &  30,000 & \\
    \rowcolor{ForestGreen!20}     \textsc{Dipole Reward} & Dipole coef value start & 0.0 & \\
    \rowcolor{ForestGreen!20}  \textsc{Coef. Schedule}   & Dipole coef iter end & 32,500 & \verb|final_entropy_iter| \\
    \rowcolor{ForestGreen!20}                            & Dipole coef value end & 2.0 & \verb|final_entropy| \\

        \bottomrule
        
    \end{tabular}
\end{table}

\subsection{Reward coefficients}

For each trained agent presented in \cref{Sec:experiments}, we used the reward coefficients listed in \cref{tbl:reward_coefs}. 
As shown in the table, we assigned a higher coefficient to validity compared to energy based reward components. This was decided based on the observed magnitude and variance of the reward components.

\begin{table}[h!]
    \centering
    \caption{Coefficients used to construct the reward functions (agents) as linear combinations of the basic reward components $\mathcal{A}$, $\mathcal{F}$ and $\mathcal{V}$. Empty corresponds to zero.}
    \label{tbl:reward_coefs}
    \begin{tabular}{c|ccc}
        \toprule 
          \textit{Agents} \textbackslash  \textit{ Components}     & $\mathcal{A}$ & $\mathcal{F}$ & $\mathcal{V}$ \\
          \midrule
          {\textbf{\color{blue}A}}     & 1 &   &   \\
          \textbf{{\color{orange}AV}}   & 1 &   & 3 \\
          \textbf{{\color{ForestGreen}F}}     &   & 1 &   \\
          \textbf{{\color{red}FV}}   &   & 1 & 3 \\
          \textbf{{\color{Plum}AFV}} & 1 & 1 & 3 \\
        \bottomrule
    \end{tabular}
\end{table}

\subsection{Hardware and compute budget}
\paragraph{Training time} All training runs were performed on a single NVIDIA RTX 3090 GPU with 8 CPU cores. A full (single seed) agent training, consisting of 20 million environment steps (incl. energy and validity metrics calculated for all generated molecules), took approximately 129 hours (5 days, 9 hours). For this paper, we trained 5 agents × 3 seeds = 15 training runs in total on the cluster.
\paragraph{Sampling time} For Q1 specifically, we sampled molecules from a single model checkpoint at a rate of 32.5 molecules per second on a NVIDIA RTX A2000 8GB Laptop GPU. For 10,000 molecules this corresponds to roughly 5 minutes per seed per agent.

\section{Metric definitions}


We split the evaluation metrics of \cref{fig:Q2} into two distinct categories. The \textit{discovery metrics} are purely count-based and contain the following metrics:

\begin{itemize}
    \item \textbf{Validity}: Validity is not directly built into the molecular generation procedure used in our framework\footnote{A word on \textit{uniqueness}: The typically reported uniqueness measure which relates the number of unique molecules to the number of sampled molecules would be a misleading metric to use in our case, since we are sampling molecules constrained to yield a pre-specified chemical formula, thereby increasing the probability of generating identical molecules compared to unconstrained sampling. As an example, we found just 3 isomers out of 10,000 generated molecules for $\text{C}_3\text{H}_8\text{O}$ in \cref{tbl:single-bag-discovery} (3 is actually the maximal number of unique molecules for this particular chemical formula). Thus, we decided to leave out this metric from \cref{fig:Q2}.}. Instead we incentivize the agent to create valid molecules based on a simple discrete reward term $r_{\text{valid}}=1$ if valid and $r_{\text{valid}}=0$ if invalid. The validity metric is straightforwardly defined as
\begin{equation}
    \text{Validity} =\frac{\# \text{valid molecules}}{\# \text{sampled molecules}}.
\end{equation}

    \item  \textbf{Rediscovery \& Expansion Ratios}:

    Relating the discovery counts to our reference dataset (QM7) helps to probe whether the agent explores broadly or if there are large gaps in its exploration. For each formula, we therefore construct the set of uniquely discovered SMILES from the RL generated molecules. Each discovered SMILES will then either be in the reference set already or correspond to a "novel" molecule, i.e. $N_{\text{unique}}^{\text{gen}} = N_{\text{rediscovered}} + N_{\text{novel}}$.
    The rediscovery and expansion ratios are calculated by relating $N_{\text{rediscovered}}$ and $N_{\text{novel}}$ to the number of reference molecules
    \begin{equation}
        \text{Rediscovery Ratio} =\frac{N_{\text{rediscovered}}}{N_{\text{unique}}^{\text{QM7}}}, \ \ \ \ \ \ \     \text{Expansion Ratio} =\frac{N_{\text{novel}}}{N_{\text{unique}}^{\text{QM7}}}.
    \end{equation}
\end{itemize}

The \textit{energy metrics} pertain to the \textit{quality} of the discovered geometries rather than their sheer \textit{quantity}. Since our agent was trained on energy based reward terms, it should be able to generate low energy isomers. However, as our PPO agent uses $3$D-spatial noise on the atomic positions in order to facilitate exploration, we must first perform structural relaxation on the generated molecules using the same \verb|xTB-GFN2| calculator that was employed for reward calculations during training. Specifically, we calculate the following energy based metrics:
\begin{itemize}
    \item \textbf{Relaxed Relative Atomic Energy (rRAE)}: This measure is defined w.r.t. our reference dataset QM7 and is calculated (at the individual molecule level) as the energy difference between our RL generated molecule and the mean energy of all the QM7 molecules of the same chemical formula (bag) \label{RAE_measure}
    \begin{equation}
        \Delta E_{\text{rRAE}}(\mathcal{C}_T) = E(\mathcal{C}_T) - \bar{E}_{\text{QM7}}^{\mathcal{B}(\mathcal{C}_T)} = E(\mathcal{C}) - \frac{1}{|\mathcal{B}(\mathcal{C}_T)|} \sum_{i=1}^{|\mathcal{B}(\mathcal{C}_T)|} E(\mathcal{C}_i^{\text{QM7}}).
    \end{equation}
    It measures the agent's joint ability to discover both low energy isomers (2D connectivity) as well as sampling low energy conformers (3D positions) for the connectivity matrix of that isomer.
    \item \textbf{Root-Mean-Square Deviation (RMSD):} To quantify how far the generated 3D structures deviate from their corresponding relaxed geometries, we compute the Root-Mean-Square Deviation (RMSD) between each generated molecule and its structure after geometry optimization using the \texttt{xTB-GFN2} method. This metric measures the average atomic displacement required to reach a local energy minimum and is defined as:
    \begin{equation}
    \text{RMSD}(\mathcal{C}_T) = \sqrt{\frac{1}{N} \sum_{i=1}^{N} \left| \mathbf{x}_i - \mathbf{x}_i^{\text{relaxed}} \right|^2},
    \end{equation}
    where $\mathbf{x}_i$ and $\mathbf{x}_i^{\text{relaxed}}$ denote the 3D positions of atom $i$ before and after end-of-episode relaxation, and $N$ is the number of atoms. A low RMSD indicates that the generated geometry was already close to a local minimum, suggesting a physically meaningful placement of atoms by the agent. In contrast, a high RMSD implies the presence of significant strain or artifacts in the initial structure that required substantial correction during optimization.
    
\end{itemize}




\section{Supplementary results of dipole finetuning (Q4)}\label{app:finetuning}
This section provides complete evaluation results for all 10 carbonate and ether formulas assessed in Q4, including the 2 formulas highlighted in the main text (H4C3O3 and H6C4O3) and the 8 formulas omitted for brevity. The finetuning setup follows the description in Section~\ref{sec:property-directed-finetuning}: the dipole moment reward coefficient was linearly ramped from 0 to 2 over 2,500 training iterations (approximately 1.28 million environment steps), and the post-finetuning evaluation checkpoint was selected prior to significant validity collapse. Note that this experiment was conducted with a single random seed, as it is intended as a proof-of-concept demonstration rather than a primary result.

\begin{figure}[h]
    \centering
    \includegraphics[width=\textwidth]{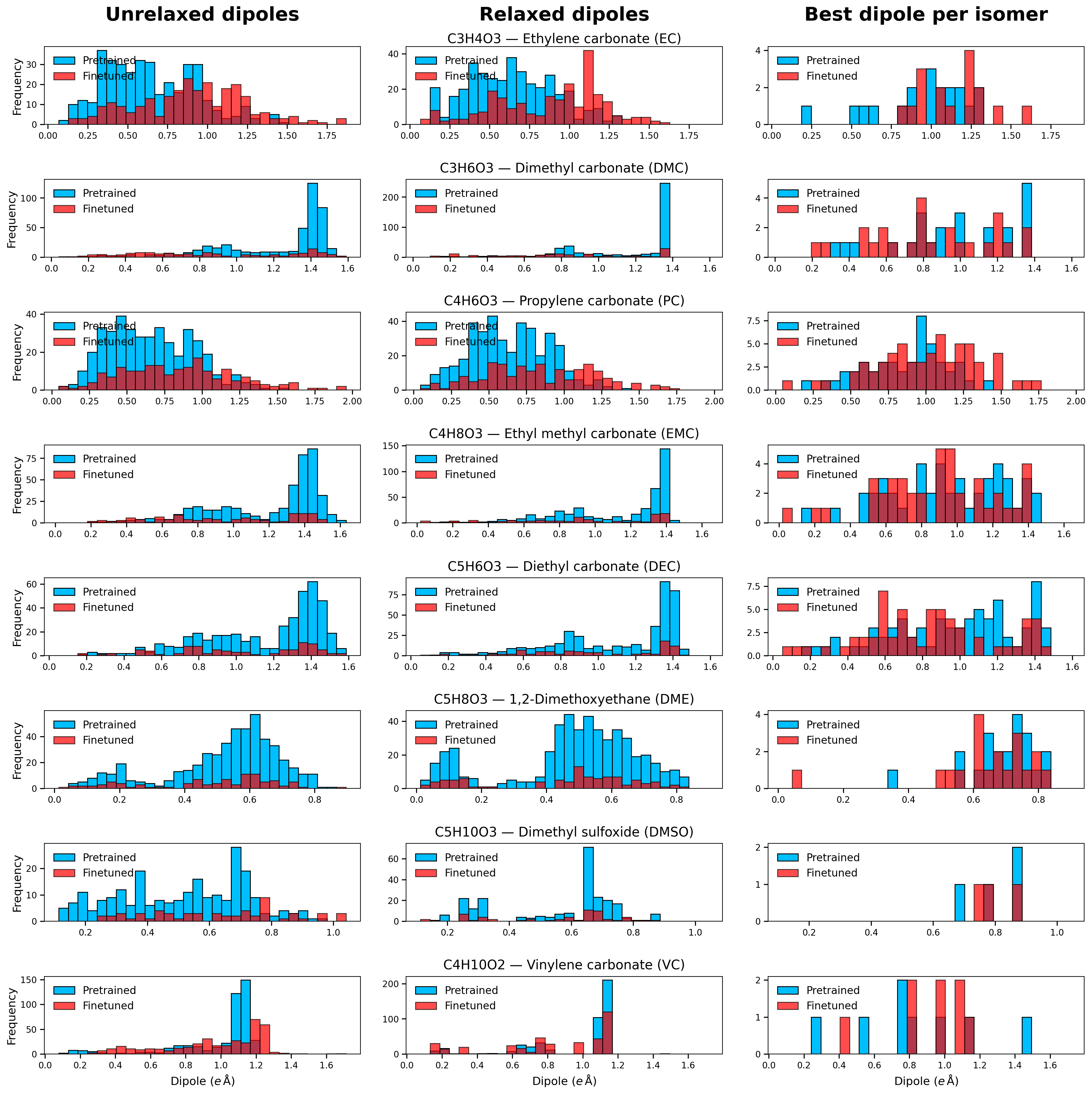}
    \caption{\textbf{Q4. Complete finetuning evaluation results.} Each row corresponds to one of the 10 evaluated carbonate and ether formulas, with the two formulas highlighted in the main text (H4C3O3 and H6C4O3) shown the first and third row. 
    For each formula, three columns are shown: (left) unrelaxed dipole moment distributions of all sampled molecules, (center) relaxed dipole moment distributions of all valid molecules, and (right) the distribution of best dipole moments per unique isomer. In each panel, the baseline agent (pre-finetuning checkpoint) is shown alongside the finetuned agent (post-finetuning checkpoint). Although the finetuned agent generates fewer valid molecules overall for several formulas due to validity collapse, the best-dipole-per-isomer column demonstrates that it retains the ability to discover comparable or improved unique isomers across most formulas.}
    \label{fig:Q4-appendix}
\end{figure}

The results are shown in \cref{fig:Q4-appendix}, where each row corresponds to one evaluation formula and each column presents a different view of the dipole moment distribution: (left) unrelaxed dipoles of all sampled molecules, (center) 
relaxed dipoles of all valid molecules, and (right) the best dipole moment achieved per unique isomer across all samples. For the 8 formulas not highlighted in the main text, the finetuned agent does not yield a meaningful improvement in 
dipole moment relative to the baseline agent. We identify two contributing factors:

\begin{itemize}
    \item \textbf{Narrow dipole moment distributions:} For several formulas, the intrinsic diversity of dipole moments accessible within the given isomer space is limited — the formula simply does not contain high-dipole structures, and finetuning therefore cannot discover what does not exist within that compositional space.
    
    \item \textbf{Reduced valid sampling:} For some formulas, the ongoing validity collapse had reduced the pool of valid molecules available for comparison. However, the best-dipole-per-isomer column (right) provides a complementary view that is robust to this effect: even when the finetuned agent generates fewer valid molecules overall, it typically retains the ability to discover comparable unique isomers. This suggests that validity collapse primarily affects sampling reliability rather than the underlying learned policy.
\end{itemize}
Together, these factors explain why the finetuning improvement is formula-dependent rather than universal. Formulas with a broader accessible dipole moment distribution and sufficient valid sampling budget are naturally more amenable to property-directed finetuning. These observations suggest that a more targeted finetuning strategy, such as selecting evaluation formulas based on the known diversity of their isomer spaces or applying a more conservative reward coefficient cap to preserve validity throughout finetuning, could yield more consistent improvements across a wider range of molecular families.

\end{document}